\documentclass[11pt]{article}

\usepackage[preprint]{acl}

\usepackage{times}
\usepackage{latexsym}
\usepackage[T1]{fontenc}
\usepackage[utf8]{inputenc}
\usepackage{microtype}
\usepackage{inconsolata}

\usepackage{amsmath}
\usepackage{amssymb}
\usepackage{booktabs}
\usepackage[table]{xcolor}
\usepackage{graphicx}
\usepackage{makecell}
\usepackage{stfloats}
\usepackage[most]{tcolorbox}
\usepackage{enumitem}

\setlist{nosep,leftmargin=*}
\definecolor{dsbg}{RGB}{235, 242, 250}
\definecolor{gptbg}{RGB}{250, 235, 235}
\colorlet{grey}{gray}

\newtcolorbox{ideabox}[2][]{
    breakable,
    enhanced,
    colback=blue!5,
    colframe=blue!40!black,
    top=0.5mm,
    bottom=0.5mm,
    boxrule=0.5mm,
    arc=2mm,
    fonttitle=\bfseries,
    fontupper=\footnotesize,
    title={\parbox{\dimexpr\linewidth-2mm\relax}{#2}},
    width=\linewidth,
    #1
}

\newtcolorbox{badbox}[2][]{
    breakable,
    enhanced,
    colback=grey!5,
    colframe=grey!40!black,
    top=0.5mm,
    bottom=0.5mm,
    boxrule=0.5mm,
    arc=2mm,
    fonttitle=\bfseries,
    title={\parbox{\dimexpr\linewidth-2mm\relax}{#2}},
    width=\linewidth,
    #1
}

\newtcolorbox{promptbox}[2][]{
    breakable,
    enhanced,
    colback=green!5,
    colframe=green!40!black,
    top=0.5mm,
    bottom=0.5mm,
    boxrule=0.5mm,
    arc=2mm,
    fonttitle=\bfseries,
    title={\parbox{\dimexpr\linewidth-2mm\relax}{#2}},
    fontupper=\small,
    width=\linewidth,
    #1
}

\title{Ideation Arena: Evaluating LLM Generated Research Ideas with Battle-style Human Expert Assessment}

\author{
  \textbf{Zhiyu Chen\textsuperscript{1,2}},
  \textbf{Keyu Zhao\textsuperscript{3}},
  \textbf{Jigao Fu\textsuperscript{4}},
  \textbf{Dong Liang\textsuperscript{4}},
  \textbf{Yanbiao Wu\textsuperscript{4}},
\\
  \textbf{Jiaoyang Li\textsuperscript{4}},
  \textbf{Haidong Xue\textsuperscript{4}},
  \textbf{Xinhua Zeng\textsuperscript{1}},
\\
  \textbf{Yuanyi Zhen\textsuperscript{2,*}},
  \textbf{Fengli Xu\textsuperscript{3,*}},
  \textbf{Yong Li\textsuperscript{3}}
\\
\\
  \textsuperscript{1}Fudan University \quad
  \textsuperscript{2}Zhongguancun Academy
\\
  \textsuperscript{3}Tsinghua University \quad
  \textsuperscript{4}Zhongguancun Institute of Artificial Intelligence
\\
  \small{
    \texttt{zhiyuchen25@m.fudan.edu.cn},
    \texttt{zhenyuanyi@bza.edu.cn},
    \texttt{fenglixu@tsinghua.edu.cn}
  }
\\[-1mm]
  \small{\textsuperscript{*}Corresponding authors.}
}

\begin{document}
\maketitle

\begin{abstract}
Evaluating research ideas generated by LLMs is difficult because their scientific value cannot be fully determined by objective criteria, and no single reference answer specifies what counts as a good idea. 
To address this challenge, we introduce Ideation Arena, a battle style platform that evaluates research ideas through pairwise human assessment. 
Ideation Arena evaluates ideas generated by 14 frontier LLMs and $5$ research agent architectures built on $2$ base models. 
To ensure a common starting point, Ideation Arena builds shared literature contexts from papers familiar to the participating researchers and provides the same contexts to all LLMs and agents. 
We collect over $6,000$ double blind pairwise comparisons from $105$ active computer science researchers and construct an Elo rating leaderboard of proposal-stage expert preferences in computer science under a shared closed-context protocol. 
We validate the rankings through interrater agreement and robustness analyses, showing that the leaderboard remains stable under changes in annotator composition and domain coverage. 
Our results show substantial variation in agent effectiveness, with some frameworks improving ideation quality over their backbones and others offering little benefit or even underperforming their base models. 
We further construct Ideation Arena Eval, a benchmark for assessing whether automated evaluators align with human preferences in research ideation. 
Experiments with current LLM judges show that they still cannot reliably reproduce expert preferences, with the best judge reaching 72.56\% Soft Accuracy on Overall Quality. 
Our code, data, and leaderboards are available at \url{https://github.com/foss12138/Research-Ideation-Arena}.
\end{abstract}

\section{Introduction}

Large Language Models (LLMs) and LLM-based agent systems are increasingly used to generate research proposals, hypotheses, and experimental plans for scientific discovery \cite{AIScientist,sciagents,reddy2025towards}. 
As these systems move from assisting with literature review to proposing new research directions, evaluating their ideation ability becomes increasingly important. 
A fluent proposal may still miss a meaningful problem, rely on weak methodological assumptions, or offer only superficial novelty. 
Despite growing interest in automated scientific discovery, the community still lacks reliable and scalable ways to assess the scientific value of research ideas generated by LLMs.

Evaluating research ideas generated by LLMs is difficult because a research idea is not defined by a single reference answer or by correctness alone \cite{doshi2024generative,liu2025researchbench}. 
Unlike code generation, which can often be checked through execution, or translation, which can be compared with semantic references \cite{tong2024codejudge,dong2025codescore,feng2025tear,qian2024large}, research ideation requires assessing whether a proposal identifies a meaningful problem, offers substantive novelty, remains feasible, and carries scientific value. 
These properties cannot be fully captured by objective metrics. 
This ambiguity also limits the reliability of automatic evaluators. 
LLM-as-a-Judge methods may favor fluent and well-structured proposals while overlooking hidden methodological flaws, weak feasibility, or limited scientific contribution \cite{canllm,kumar2025can}. 
For open-ended tasks such as research ideation, where no standard answer defines success, evaluation therefore requires expert involvement. 
Domain researchers are better positioned to judge whether an idea is genuinely novel, methodologically feasible, scientifically significant, and scientifically valuable.

Pairwise human preference evaluation offers a natural way to organize such expert judgments, since it allows evaluators to compare two candidate ideas without requiring a single gold reference. 
Arena style platforms such as Chatbot Arena \cite{chatbot} have shown that pairwise voting can scale human evaluation and support preference-based leaderboards. 
However, existing arena platforms are designed mainly for general-purpose assistants and broad user preferences, rather than expert assessment of scientific ideas.

The central challenge is therefore to build a trustworthy and scalable human-in-the-loop evaluation infrastructure for research ideation: one that enables different LLMs and agent systems to generate ideas under comparable conditions, supports expert assessment with consistent criteria, and aggregates subjective judgments into robust model-level comparisons.

\begin{figure*}[htbp]
    \centering
    \includegraphics[width=\linewidth]{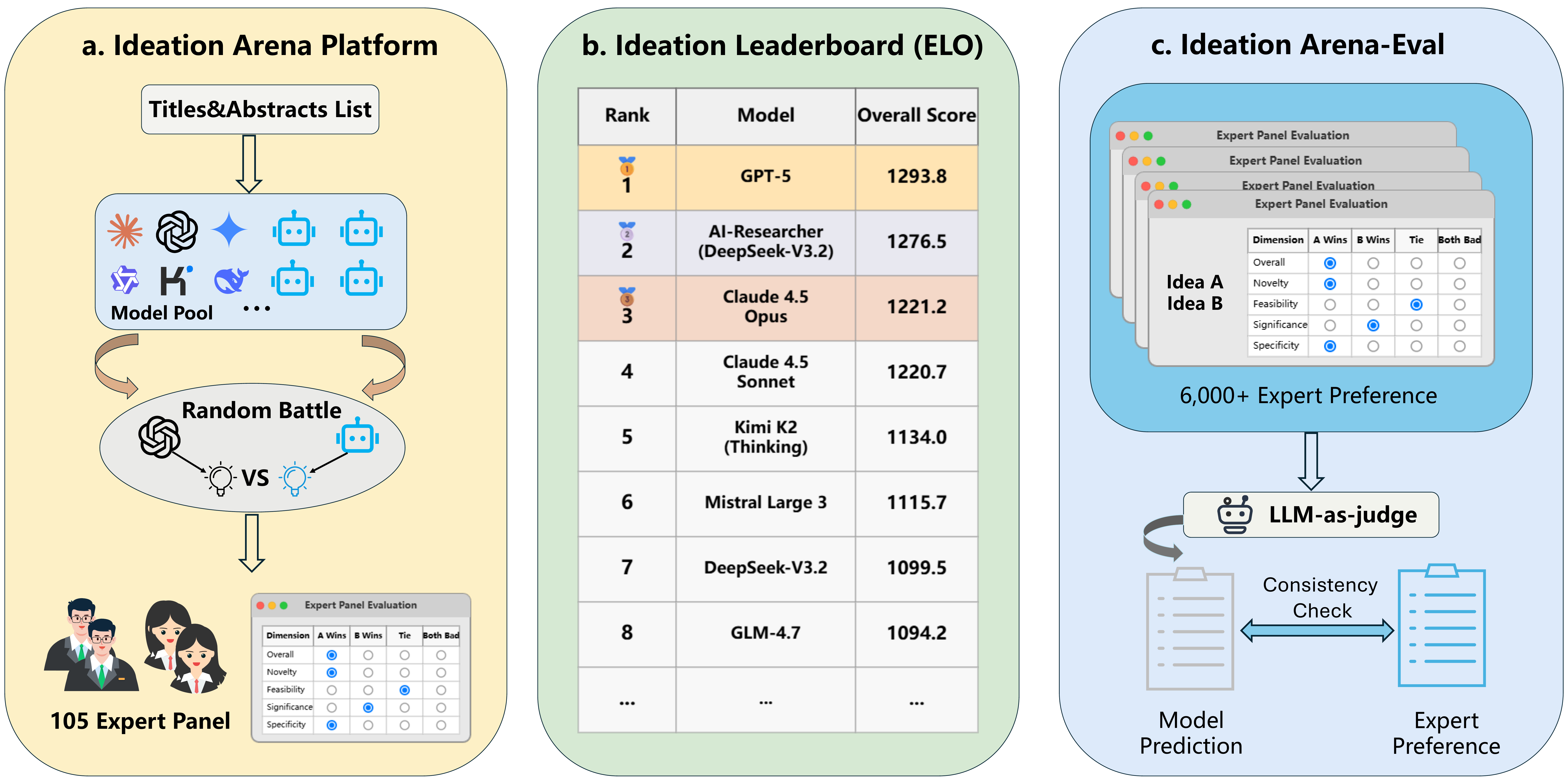 } 
    \vspace{1mm}
    \caption{Overview of Ideation Arena, a battle style platform for evaluating research ideas. The framework comprises (a) the Ideation Arena Platform, which addresses open-ended evaluation challenges through shared literature contexts and double blind pairwise comparisons, (b) the Ideation Leaderboard, which utilizes Elo-based rankings to quantify the scientific reasoning capabilities of diverse models and agent architectures, and (c) Ideation Arena-Eval, the first meta-evaluation benchmark designed to assess the alignment between automated evaluators and human expert preferences.}
    \vspace{1mm}
    \label{fig:intro-case}
\end{figure*}

To address this challenge, we introduce \textbf{Ideation Arena}, which operationalizes research idea assessment as battle style evaluation, as shown in Figure \ref{fig:intro-case}. 
Ideation Arena uses double blind pairwise comparisons by active researchers to aggregate open-ended assessments into a preference-based leaderboard. 
The shared-context design controls the information available to each system by grounding every comparison in the same literature context selected from papers familiar to the participating researchers. 
This design keeps the evaluation focused on idea generation rather than differences in retrieval results, tool use, or external background knowledge.

We evaluate ideas generated by 14 frontier LLMs and $5$ research agent architectures built on $2$ base models. 
In total, $105$ active computer science researchers provide over $6,000$ double blind pairwise comparisons across Overall Quality, Novelty, Feasibility, Significance, and Specificity. 
Based on these comparisons, we construct an Elo rating leaderboard using the Bradley-Terry model. 
Reliability analyses further show that the rankings are not driven by a small subset of annotators or domains. 
The resulting leaderboard identifies GPT-5.1, AI-Researcher with DeepSeek V3.2, Claude 4.5 Opus, and Claude 4.5 Sonnet as the strongest overall performers, while also showing that agent gains are highly architecture dependent. 
For example, DeepSeek V3.2 ranks 7th as a base model with an Overall Quality score of $1099.5$, whereas AI-Researcher with DeepSeek V3.2 ranks 2nd with a score of $1276.5$. 
At the same time, all other agent architectures built on DeepSeek V3.2 score below the standalone DeepSeek V3.2 baseline. 
This contrast suggests that the way an agent workflow organizes reasoning, reflection, and proposal construction can substantially shape ideation quality.

To examine whether the evaluation protocol extends beyond computer science, we further conduct pilot studies in Biology and Physics. 
Domain experts in both fields annotate model-generated ideas with the same double blind pairwise comparison procedure, and the resulting trends are broadly consistent with the main benchmark.

We further construct \textbf{Ideation Arena Eval}, a benchmark for assessing whether automated evaluators align with human preferences in research ideation. 
Experiments with 14 frontier LLM judges show that current models still struggle to match human preferences: most models score below 70\% agreement across dimensions, and the best model reaches only $72.56\%$ agreement on Overall Quality. 
The gap is especially clear in Feasibility, suggesting that LLM judges have difficulty identifying ideas that appear plausible but are methodologically weak. 
These findings highlight the need for human grounded evaluation when measuring progress in automated scientific discovery.

Our contributions are summarized as follows:

\begin{itemize}
    \item We introduce \textbf{Ideation Arena}, an open battle style platform for evaluating LLM-generated research ideas through double blind pairwise human comparisons from active researchers.

    \item We present the \textbf{Research Ideation Leaderboard}, derived from over 6,000 expert pairwise comparisons across five evaluation dimensions. We validate ranking reliability through interrater agreement and robustness analyses, and show that agent architectures exhibit highly variable effects on base-model ideation quality.

    \item We develop \textbf{Ideation Arena-Eval}, a meta-evaluation benchmark for auditing automated research idea evaluators. Experiments with 14 foundation models reveal persistent misalignment between LLM judges and human preferences, especially for feasibility-oriented judgments.
\end{itemize}

\section{Related Work}

LLMs are increasingly extending beyond general-purpose conversational applications to specialized academic and scientific tasks, including their use as autonomous agents for scientific discovery\cite{stephanie2,moose-chem,virsci,lu2024ai}. Systems such as AI-Researcher\cite{AI-Researcher}, ResearchAgent\cite{researchagent}, and SciMON\cite{scimon} can generate research ideas end to end, yet the community still lacks reliable standards for judging whether these ideas contain methodological depth comparable to human expert proposals. Unlike code generation or mathematical problem solving, research ideation is open-ended and has no single ground truth. Existing evaluation protocols therefore often conflate fluent presentation with genuine scientific value\cite{weidinger2025toward,sottana2023evaluation}.

Several automated benchmarks have been proposed to address this gap. AI Idea Bench 2025\cite{qiu2025ai} compares generated ideas with real papers and reference materials, but its emphasis on target-paper alignment is closer to reconstructing a known idea than evaluating divergent ideation. IdeaBench\cite{ideabench} and LiveIdeaBench\cite{liveideabench} broaden the task forms and scoring dimensions, including novelty and feasibility, but still rely heavily on LLM-as-a-Judge. Such evaluators may miss hallucinated innovations that appear coherent while lacking technical feasibility, especially when domain expertise is required.

Human preference evaluation offers a complementary path. MT-Bench and Chatbot Arena establish LLM-as-a-Judge benchmarking and crowdsourced pairwise preference evaluation for general-purpose assistant responses\cite{mtbench,chatbot}. Auto-Arena automates pairwise evaluation through agent peer battles and committee discussions\cite{autoarena}. The arena paradigm has also been extended to vision, search, generation, forecasting, and embodied tasks\cite{visionarena,genai,miroyan2025search,yang2025llm,ni2025embodied}. However, these platforms mainly assess perception, execution, or known-information retrieval rather than open-ended scientific ideation. SciArena\cite{sciarena} targets scientific literature processing but focuses on knowledge synthesis, while \citet{canllm} compare LLM-generated and human ideas through a static double-blind study. In contrast, \textbf{Ideation Arena} provides an open, expert-driven arena for research ideation, aligns model inputs with evaluator expertise through citation-based data construction, and uses the resulting human preferences to build \textbf{Ideation Arena-Eval} for auditing automated idea evaluators.

\section{Data Construction: An Expert-Guided Retrospective Pipeline}

\begin{figure*}[t] % 跨栏图片通常建议放在页面顶部 [t]
   \centering
   \includegraphics[width=0.8\textwidth]{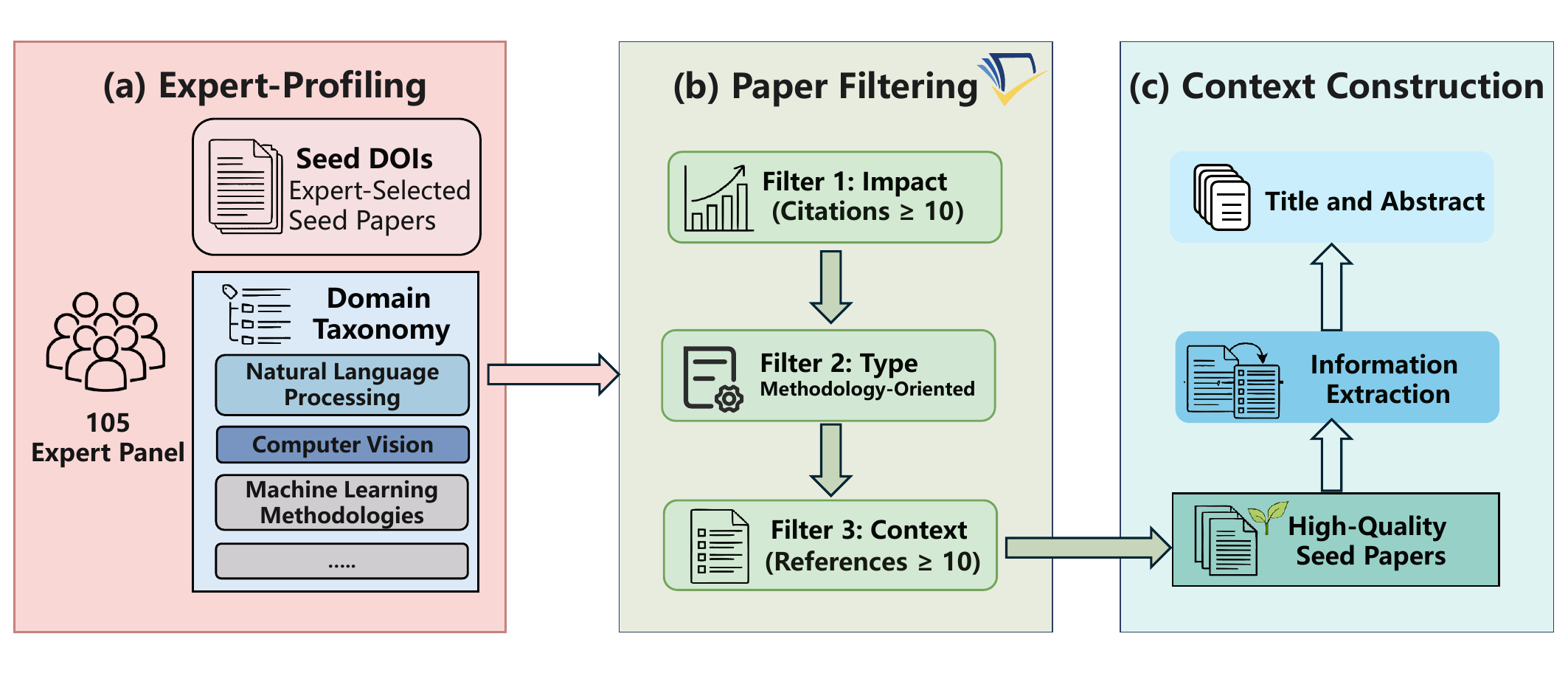} 
   \caption{Illustration of the Expert-Guided Retrospective Data Construction Pipeline.}
   \label{data_pipline}
\end{figure*}

To establish a high-quality evaluation benchmark, we devised an expert-guided data construction pipeline (Figure~\ref{data_pipline}). By leveraging the prior knowledge of domain experts to guide data acquisition, we construct standardized model inputs.

\subsection{Standardized Task Definition}

In contrast to traditional chat-based arenas that rely on dynamic user inputs~\cite{chatbot,sciarena,visionarena}, Ideation Arena establishes a Retrospective Ideation Paradigm employing preset queries to ensure evaluation consistency. Rather than responding to open-ended user instructions, the system provides models with background citations, including titles and abstracts, regarding a specific research problem, requiring the generation of substantively innovative methodological proposals. This design simulates the authentic cognitive process in which researchers identify unresolved issues and formulate potential solutions after reviewing the literature. Furthermore, this framework eliminates the variance introduced by prompt engineering, ensuring that the evaluation focuses strictly on the intrinsic capability of research ideation rather than on conversational skills tailored to user preferences.

\subsection{Data Acquisition and Filtering Pipeline}

We constructed granular research profiles for each participant. Specifically, participants were required to submit DOIs of papers they authored or recently studied to establish a user background knowledge base. Concurrently, drawing upon the AAAI submission subject guidelines, we established a computer science taxonomy for domain profiling; the detailed category list is provided in Appendix~\ref{sec:appendix_taxonomy}. Participants were invited to select their fields of primary interest to define their academic domain profiles.

Leveraging these DOIs and field selections, we performed targeted retrieval and random sampling of papers published within the last five years via the Semantic Scholar API\cite{s2} to construct an initial candidate paper pool. To ensure the scientific value and feasibility of this hybrid corpus, we designed a strict three-level filtering mechanism. First, we applied an impact filter, retaining only papers with at least 10 citations to ensure their recognition and baseline quality within the academic community. Second, we executed a type filter based on metadata, excluding surveys, benchmarks, and pure evaluation papers to strictly preserve ``methodology-oriented'' research work characterized by concrete technical contributions. Finally, to guarantee context richness, we implemented a context completeness filter, requiring a minimum of 10 references per paper. This threshold aligns with empirical settings in prior work~\cite{ideabench,canllm}, providing sufficient information density to ground logical reasoning and stimulate valid research ideation. The final pool contains 2,191 unique seed-paper contexts spanning all eight primary areas; Appendix~\ref{sec:appendix_seed_distribution} reports the primary-area and leading secondary-subfield distributions.

\subsection{Problem Formalization}

We mathematically formalize the research ideation task as a conditional generation problem within a constrained information environment. Let $\mathcal{P}_{seed}$ denote a target research paper. We aim to simulate the discovery context of the authors prior to the composition of $\mathcal{P}_{seed}$ by reconstructing the information exclusively from the prior works cited in the paper. For a given seed paper, we extract the reference list $\mathcal{R} = \{r_1, r_2, \dots, r_N\}$. To construct the input context, we concatenate the title ($T$) and abstract ($A$) of each reference, yielding a standardized query context $C = \bigoplus_{i=1}^{N} \left( \text{Title}(r_i) \oplus \text{Abstract}(r_i) \right)$ that represents the boundary of observable knowledge.

The context $C$ is integrated into a comprehensive task instruction $I$, designed to guide the model in identifying research gaps within $C$ and proposing a novel methodology. The input query $Q$ provided to the LLM or scientific agent is formally defined as the tuple $(I, C)$. Let $\mathcal{M}$ denote the foundation model or scientific agent, and the generation process is formalized as $\hat{P}_{model} = \mathcal{M}(Q) = \mathcal{M}(I, C)$.
Crucially, we impose a Closed-Context Constraint: the model $\mathcal{M}$ is instructed to derive insights exclusively from $C$, strictly prohibiting access to external knowledge bases or internet retrieval. This requirement ensures that the generated idea $\hat{P}_{model}$ originates solely from reasoning over the provided literature.

\section{The Ideation Arena Platform}

This section introduces the Ideation Arena infrastructure, covering its closed-context model pool, double-blind evaluation protocol, and Elo-based ranking formulation.

\subsection{Platform Architecture and Evaluation Protocol}

We curated a representative model pool comprising $8$ proprietary LLMs, $6$ frontier open-source models, and $5$ specialized research agents. To decouple agent architecture from backbone capability, each research agent was evaluated in two variants instantiated with GPT-4o and DeepSeek V3.2.

We adapted each research agent to the shared closed-context protocol by retaining its reasoning, iterative-refinement, and proposal-generation components while disabling external retrieval and pre-constructed knowledge sources. All agents and foundation models receive the same citation titles and abstracts as their available literature context. Appendix~\ref{sec:appendix_closed_context} lists the retained, removed, and replaced components for each agent.

For each seed task, every system generates a proposal from the same closed-context citation information. We randomly sample proposals from two distinct systems within that task, randomly assign them to positions A and B, and display the pair without system identities. Each battle is assigned to an expert whose self-reported expertise matches the seed-paper topic. Experts rate each pair on Novelty, Feasibility, Significance, Specificity, and Overall Quality, assigning one of four verdicts: ``A Wins,'' ``B Wins,'' ``Tie,'' or ``Both Bad.'' Annotators assess Novelty using the provided literature context and their domain expertise, completing the judgment without consulting external papers or web resources. Detailed criteria are provided in Appendix~\ref{sec:appendix_eval_criteria}.

\subsection{ELO-Based Ranking System}

To translate discrete, sparse expert pairwise votes into a global metric, we employ a rating system based on the Bradley-Terry model\cite{btmodel} for ELO score estimation\cite{elo}.

We parameterize the Research Ideation Capability of model $m_i$ as a latent coefficient $s_i \in \mathbb{R}$. In a pairwise comparison, assuming model $m_i$ competes against $m_j$, the probability $P(i \succ j)$ that $m_i$ defeats $m_j$ is determined by the logistic function of the difference in their capability coefficients:
\begin{equation}
 P(i \succ j \mid s_i, s_j) = \frac{1}{1 + e^{-(s_i - s_j)}} 
\end{equation}
We treat both ``Tie'' and ``Both Bad'' as draws, setting $y_{ij}=0.5$. We denote the entire Arena dataset as $\mathcal{D} = \{(i, j, y_{ij})_k\}_{k=1}^N$, where $y_{ij} \in \{0, 0.5, 1\}$ represents loss, tie, and win states, respectively. Appendix~\ref{sec:appendix_bothbad} reports a sensitivity analysis that models ``Both Bad'' separately and reports its model-level involvement rates.

We formulate the parameter estimation as a Maximum Likelihood Estimation (MLE) problem. By maximizing the log-likelihood function of the observed data, we solve for the optimal capability coefficients $\hat{\mathbf{s}}$:
\begin{equation}
\begin{aligned}
 \hat{\mathbf{s}} = \operatorname*{argmax}_{\mathbf{s}} \sum_{k=1}^{N}
 \Big[& y_{ij}^{(k)} \ln P(i \succ j) \\
 &+ (1 - y_{ij}^{(k)}) \ln P(j \succ i) \Big]
\end{aligned}
\end{equation}
The resulting estimates $\hat{\mathbf{s}}$ are scaled and converted into standard Elo ratings to construct the leaderboard. Given the sparsity of pairwise comparison data, a single point estimate cannot fully quantify ranking uncertainty. We employ a Bootstrap method with 1,000 resamples to calculate the 95\% confidence interval for each score, thereby assessing the statistical significance of performance differences between models.
\begin{table}[h]
\centering
\caption{\textbf{Inter-Annotator Agreement (IAA) Analysis.}}
\label{tab:iaa}
\resizebox{\columnwidth}{!}{%
\begin{tabular}{lcc}
\hline
\textbf{Dimension} & \textbf{Agreement Accuracy} & \textbf{Krippendorff's $\alpha$} \\
\hline
Overall Quality & 56.8\% & 0.6088 \\
Novelty         & 54.4\% & 0.5884 \\
Feasibility     & 53.8\% & 0.5863 \\
Significance    & 56.5\% & 0.5963 \\
Specificity     & 55.9\% & 0.5996 \\
\hline
\end{tabular}%
}
\end{table}

\begin{table*}[t]
    \centering
    \small
    \caption{Leaderboard of AI Agents under strict closed-context constraints. All agents were evaluated with online retrieval disabled and reasoning restricted solely to the provided citation contexts. The best results are highlighted in \textbf{bold}, and the second-best results are \underline{underlined}. Rows highlighted in \colorbox{dsbg}{\textbf{light blue}} denote agents based on \textbf{DeepSeek V3.2}, and rows in \colorbox{gptbg}{\textbf{light red}} denote agents based on \textbf{GPT-4o}.}
    \label{tab:ideation_arena_leaderboard}
    
    \resizebox{\textwidth}{!}{%
    \begin{tabular}{l l c c c c c c c}
        \toprule
        \textbf{Rank} & \textbf{Agent Name} & \textbf{Battles} & \makecell[c]{\textbf{Tokens/} \\ \textbf{Query} $\downarrow$} & \makecell[c]{\textbf{Overall} \\ \textbf{Quality} $\uparrow$} & \textbf{Novelty} $\uparrow$ & \textbf{Feasibility} $\uparrow$ & \textbf{Significance} $\uparrow$ & \textbf{Specificity} $\uparrow$ \\
        \midrule
        1 & \textbf{GPT-5.1} & 542 & 23377.1 & \textbf{1293.8} & \textbf{1208.6} & \textbf{1180.9} & \textbf{1236.8} & \textbf{1275.3} \\

        \rowcolor{dsbg}
        2 & \textbf{AI-Researcher (DeepSeek V3.2)} & 489 & 99067.8 & \underline{1276.5} & \underline{1203.4} & \underline{1167.5} & \underline{1236.3} & 1232.0 \\
        
        3 & \textbf{Claude 4.5 Opus} & 643 & 16737.3 & 1221.2 & 1141.3 & 1160.2 & 1149.2 & \underline{1240.3} \\
        
        4 & \textbf{Claude 4.5 Sonnet} & 597 & 17183.7 & 1220.7 & 1131.2 & 1163.6 & 1146.3 & 1214.7 \\
        
        5 & \textbf{Kimi K2 (Thinking)} & 610 & 19945.7 & 1134.0 & 1132.8 & 1060.7 & 1103.2 & 1118.6 \\
        
        6 & \textbf{Mistral Large 3} & 575 & 15385.3 & 1115.7 & 1081.2 & 1102.2 & 1091.4 & 1106.3 \\

        \rowcolor{dsbg}
        7 & \textbf{DeepSeek V3.2} & 606 & 14860.3 & 1099.5 & 1081.8 & 1073.9 & 1089.1 & 1071.3 \\
        
        8 & \textbf{GLM-4.7} & 626 & 22364.6 & 1094.2 & 1080.7 & 1049.7 & 1058.7 & 1059.9 \\
        
        9 & \textbf{OpenAI o3} & 547 & 16848.0 & 1083.6 & 1064.6 & 1050.0 & 1051.7 & 1059.6 \\
        
        10 & \textbf{Gemini 3 Pro Preview} & 495 & 18294.1 & 1070.3 & 1082.3 & 1020.1 & 1049.5 & 1031.5 \\
        
        11 & \textbf{Gemini 3 Flash Preview} & 619 & \underline{14323.9} & 1069.2 & 1088.4 & 1038.5 & 1038.0 & 1055.3 \\
        
        12 & \textbf{Grok 4 Fast} & 625 & 15470.9 & 1029.3 & 1010.0 & 1025.1 & 1002.1 & 1049.8 \\

        \rowcolor{gptbg}
        13 & \textbf{AI-Researcher (GPT-4o)} & 534 & 74575.2 & 971.8 & 979.6 & 984.8 & 1001.7 & 956.5 \\
        
        14 & \textbf{Grok 4} & 561 & 15628.9 & 951.5 & 967.7 & 983.1 & 958.9 & 962.8 \\

        \rowcolor{dsbg}
        15 & \textbf{Virtual Scientists (DeepSeek V3.2)} & 455 & 48531.8 & 930.0 & 977.6 & 924.4 & 963.7 & 967.2 \\
        
        16 & \textbf{Qwen3-Max} & 475 & 14704.8 & 929.0 & 935.6 & 993.2 & 929.5 & 950.2 \\

        \rowcolor{dsbg}
        17 & \textbf{ResearchAgent (DeepSeek V3.2)} & 512 & 119009.0 & 910.6 & 906.0 & 934.8 & 931.2 & 936.1 \\
        
        \rowcolor{dsbg}
        18 & \textbf{SciMON (DeepSeek V3.2)} & 516 & 32595.3 & 871.7 & 955.3 & 892.8 & 940.3 & 869.6 \\

        \rowcolor{gptbg}
        19 & \textbf{Virtual Scientists (GPT-4o)} & 460 & 33347.8 & 843.2 & 863.4 & 922.0 & 868.5 & 882.1 \\
        
        \rowcolor{gptbg}
        20 & \textbf{ResearchAgent (GPT-4o)} & 494 & 88297.0 & 838.9 & 834.0 & 903.2 & 872.5 & 835.6 \\
        
        \rowcolor{gptbg}
        21 & \textbf{GPT-4o} & 434 & \textbf{13979.8} & 829.9 & 881.3 & 925.0 & 866.9 & 836.2 \\

        \rowcolor{dsbg}
        22 & \textbf{MOOSE-Chem (DeepSeek V3.2)} & 487 & 104,535.1 & 820.9 & 853.4 & 859.3 & 859.7 & 856.4 \\

        \rowcolor{gptbg}
        23 & \textbf{SciMON (GPT-4o)} & 506 & 32494.7 & 728.1 & 815.2 & 813.4 & 817.3 & 714.1 \\
        
        \rowcolor{gptbg}
        24 & \textbf{MOOSE-Chem (GPT-4o)} & 474 & 77,226.7 & 675.2 & 728.6 & 774.2 & 740.0 & 719.5 \\
        \bottomrule
    \end{tabular}%
    }
\end{table*}
\subsection{Reliability of Expert Evaluation}

In total, we collected over 6,000 valid pairwise votes. The 105-person panel includes PhD students, postdoctoral researchers, faculty members, and industry researchers, with self-reported expertise spanning all eight primary areas; Appendix~\ref{sec:appendix_expert_panel} reports the area-level composition. To assess the reliability of expert evaluations, we conducted an inter-annotator agreement analysis on comparisons that received redundant annotations. Due to domain overlaps in our assignment mechanism, 1,675 queries, corresponding to 3,888 pairwise battles, were independently reviewed by at least two Expert Panel members. On this subset, we measured inter-annotator agreement using Pairwise Agreement Accuracy and Krippendorff’s Alpha, treating ties as half votes for both sides.
As shown in Table~\ref{tab:iaa}, Krippendorff's $\alpha$ remains consistently around 0.60 across all five dimensions, with Pairwise Agreement Accuracy reaching 56.8\% for Overall Quality. These results are comparable to prior expert evaluations of open-ended research ideas~\cite{canllm}, suggesting meaningful consensus among Expert Panel members.

Beyond annotation-level agreement, we further test the stability of the final leaderboard under perturbations of the evaluator pool and domain coverage. The re-estimated leaderboards remain highly consistent with the full leaderboard under both 50\% annotator subsampling over 200 runs and leave-one-domain-out re-estimation, with average Spearman correlations of 0.980 and 0.998, respectively. This suggests that our benchmark-level conclusions are robust to annotator composition and domain-specific preference variation.

\section{Ideation Arena Analysis}

In this section, we provide an in-depth analysis of the \textbf{Ideation Arena} leaderboard results and investigate the performance disparities across various model architectures within the domain of research idea generation.

\begin{table}[!b]
\centering
\small
\caption{Preliminary Biology and Physics pilot results. We report Bradley--Terry ratings on Overall Quality.}
\label{tab:cross_domain}
\begin{tabular}{lcc}
\toprule
Model & Biology & Physics \\
\midrule
GPT-5.1 & 1237.6 & 1375.6 \\
AI-Researcher (DeepSeek V3.2) & 1165.5 & 1106.5 \\
DeepSeek V3.2 & 854.8 & 916.1 \\
GPT-4o & 742.1 & 601.8 \\
\bottomrule
\end{tabular}
\end{table}

% Enqueue the wide meta-evaluation table early enough to place it before the
% final main-text page while preserving its numbering after Table~\ref{tab:cross_domain}.
\begin{table*}[t!]
    \centering
    \small
    \caption{Alignment of LLM Judges with Human Expert Votes on Ideation Arena-Eval. Accuracy is reported with ties calculated as 0.5 votes. The \textit{Random Guess} baseline reflects the expected score based on the ground truth distribution (see Appendix \ref{sec:appendix_tie_rates} for details). Best results are in \textbf{bold}, second best are \underline{underlined}.}
    \label{tab:meta_eval}
    \resizebox{\textwidth}{!}{%
    \begin{tabular}{lccccc}
    \hline
    \textbf{Model} & \textbf{Overall Quality} $\uparrow$ & \textbf{Novelty} $\uparrow$ & \textbf{Feasibility} $\uparrow$ & \textbf{Significance} $\uparrow$ & \textbf{Specificity} $\uparrow$ \\ \hline
    Random Guess & 50.57\% & 51.29\% & 51.68\% & 51.85\% & 51.14\% \\ \hline
    Gemini 3 Pro Preview & \textbf{72.56\%} & \textbf{68.48\%} & \textbf{64.56\%} & \textbf{65.68\%} & \underline{67.60\%} \\
    Claude 4.5 Sonnet & \underline{68.78\%} & \underline{65.96\%} & 59.98\% & 61.97\% & \textbf{69.25\%} \\
    Grok 4 & 65.81\% & 62.81\% & 60.34\% & \underline{63.05\%} & 65.17\% \\
    Kimi K2 (Thinking) & 65.13\% & 64.26\% & 57.91\% & 62.55\% & 64.44\% \\
    GLM-4.7 & 64.76\% & 63.35\% & 58.73\% & 62.46\% & 64.98\% \\
    Grok 4 Fast & 64.42\% & 61.72\% & \underline{60.36\%} & 61.61\% & 64.49\% \\
    Claude 4.5 Opus & 64.64\% & 62.47\% & 57.19\% & 62.64\% & 64.58\% \\
    Qwen3-Max & 65.50\% & 62.74\% & 57.47\% & 61.45\% & 63.07\% \\
    GPT-5.1 & 64.29\% & 61.80\% & 57.74\% & 60.70\% & 65.30\% \\
    DeepSeek V3.2 & 63.30\% & 62.22\% & 57.85\% & 61.64\% & 64.14\% \\
    Mistral Large 2512 & 62.83\% & 62.48\% & 58.64\% & 62.61\% & 62.31\% \\
    Gemini 3 Flash Preview & 63.62\% & 62.53\% & 56.07\% & 60.81\% & 63.24\% \\
    OpenAI o3 & 62.69\% & 61.66\% & 54.88\% & 59.58\% & 63.87\% \\
    GPT-4o & 59.52\% & 59.95\% & 55.72\% & 56.30\% & 58.78\% \\ \hline
    \end{tabular}%
    }
\end{table*}

\subsection{Overall Leaderboard Performance}
Table \ref{tab:ideation_arena_leaderboard} presents the overall results of the Ideation Arena leaderboard. Detailed confidence intervals are provided in the Appendix \ref{sec:appendix_stats}. In terms of overall ranking, model performance exhibits significant stratification. GPT-5.1, the AI-Researcher (DeepSeek V3.2) agent, Claude 4.5 Opus, and Claude 4.5 Sonnet demonstrate a distinct lead over other models, all scoring above $1200$. Notably, GPT-5.1 secured the first or second position across all sub-dimensions, demonstrating superior general reasoning capabilities. Kimi K2 (Thinking) ranks fifth overall, emerging as the top-performing open-source model. A critical insight lies in the performance delta introduced by agent frameworks. While DeepSeek V3.2 ranks 7th ($1099.5$) as a standalone base model, its encapsulation within the AI-Researcher framework propels it to 2nd place ($1276.5$). This substantial increase of $+177$ points indicates that well-designed reflexive workflows can effectively elicit latent reasoning capabilities in open-source models, enabling them to compete with proprietary state-of-the-art systems. Concurrently, due to the rapid iteration of foundation models, DeepSeek V3.2 outperforms GPT-4o in base model rankings. This advantage persists when both are encapsulated within the same agent framework, suggesting that the upper bound of an agent's capability is constrained by the native capacity of its underlying base model.

We further analyze the average token consumption per query for each model to evaluate the computational efficiency of research ideation. The data distribution reveals that token consumption for agent systems is generally significantly higher than that of foundation models, presenting two distinct efficacy patterns. GPT-5.1 and Claude 4.5 exhibit superior ``Parametric Efficiency,'' achieving state-of-the-art performance with lower token overhead ($\sim 23$k). This indicates that stronger base models can generate high-density insights without relying on extensive external frameworks. Conversely, AI-Researcher equipped with DeepSeek V3.2 consumes $7\times$ the tokens of its base model to achieve a higher ranking, demonstrating that performance gains can be effectively achieved by increasing inference-time compute through iterative reflection. However, mere scaling does not guarantee quality. Agents such as ResearchAgent (DeepSeek V3.2) incur the highest computational cost ($119$k tokens) but rank significantly lower ($17$th). This discrepancy suggests that expert-preferred ideation depends less on the amount of agentic interaction and more on whether the workflow produces structured, specific, and methodologically grounded proposals. We provide a detailed discussion of closed-context effects on agent performance in Appendix~\ref{sec:appendix_closed_context}.

To further characterize expert preference patterns, we conducted a correlation analysis of the Elo ratings across evaluation dimensions on the leaderboard. We observed a universally high correlation across all metrics ($r > 0.95$), suggesting that the research ideation capability of current LLMs is driven by a holistic ``general reasoning capability'' rather than by isolated specialized skills. Within this overall pattern, Overall Quality is especially aligned with Specificity, suggesting that expert judgments of overall merit are closely tied to the concreteness and methodological detail of a proposal.

Because experts compare complete proposal texts, we analyze response length as a measured presentation factor. Among battles between systems whose anchored ratings differ by at most 100 points, the longer proposal is preferred in 52.8\%--58.7\% of comparisons across the five dimensions. Length-adjusted Bradley--Terry rankings have Spearman correlations of 0.7504--0.8835 with the original rankings; Appendix~\ref{sec:appendix_length} gives the model, dimension-level results, and ranking-sensitivity analysis.

We also examine whether leaderboard position is associated with similarity to the corresponding seed paper. After normalizing SPECTER2 similarity by the mean similarity among proposals generated for the same task, seed-proximity margins have model-level Spearman correlations of $-0.041$ with Overall Quality Elo and $-0.082$ with Novelty Elo. Appendix~\ref{sec:appendix_seed_proximity} reports the construction, ranking-group means, and confidence intervals.

\subsection{Cross-Domain Evaluation Pilot}

We conduct preliminary applications of the Ideation Arena protocol in Biology and Physics. Each pilot includes four domain experts, with each expert completing approximately 50 pairwise comparisons and overlapping annotations used to estimate agreement. Mean pairwise agreement is 0.539 in Biology and 0.556 in Physics. Table~\ref{tab:cross_domain} reports Bradley--Terry ratings on Overall Quality for four systems; Appendix~\ref{sec:appendix_cross_domain} reports dimension-level agreement and bootstrap confidence intervals.

\subsection{Case Study Analysis}
We examined sample responses generated by high-ranking models (including GPT-5.1 and AI-Researcher utilizing DeepSeek V3.2) and low-ranking counterparts (such as SciMON with GPT-4o and MOOSE-Chem with GPT-4o). Our analysis reveals that high-performing models produce comprehensive, structured academic proposals that incorporate specific mathematical formulations, detailed algorithms, step-by-step methodological derivations, and concrete experimental plans. In contrast, low-ranking models demonstrate a marked deficiency in informational depth. Through inductive analysis, we classify their limitations into three primary categories: \textbf{(1) Brevity}, where models provide only high-level overviews resembling abstracts rather than full proposals, \textbf{(2) Conceptual Hollowness}, characterized by the use of generic terminology without specific mathematical definitions or architectural details, and \textbf{(3) Structural Deficiency}, which involves the absence of problem analysis, resulting in incomplete proposal formulations.
Comprehensive examples of each category are provided in the Appendix \ref{sec:appendix_case_study}. These findings suggest that human evaluators prioritize substantive completeness during blind reviews. In the ideation phase, which lacks mechanisms for external verification, the formal depth and logical consistency of an idea significantly influence the assessment of scientific value by reviewers.

% \begin{figure}[t]
%     \centering
%     \includegraphics[width=\linewidth]{figure/prompt.pdf}
%     \caption{The standardized prompt template used for the LLM-as-a-Judge component in Ideation Arena-Eval. Models perform pairwise comparisons of research proposals across five dimensions.}
%     \label{fig:prompt}
% \end{figure}

\section{Meta-Evaluation: The Ideation Arena-Eval Benchmark}

Given the prevalence of LLM-as-a-Judge approaches in evaluating research ideation, assessing the reliability of LLMs in this specific context is critical. To this end, we construct Ideation Arena-Eval, a benchmark derived from our expert-annotated corpus. This benchmark comprises over 6,000 idea pairs annotated with expert preferences across five distinct dimensions.

\subsection{Experimental Setup and Metrics}

We formulate the evaluation as a ternary classification task (win, loss, tie), employing the standardized prompt template in Appendix \ref{sec:prompt}. To accommodate the inherent ambiguity of research ideation, we employ a Soft-Accuracy metric, wherein a tie in the ground truth equitably assigns a score of $0.5$ to both candidates to ensure impartiality. For rigorous benchmarking, we establish a random guessing baseline derived specifically from the marginal distribution of labels in the test set. Because ties receive partial credit and their frequencies vary across evaluation dimensions, the expected accuracy ranges from $50.57\%$ to $51.85\%$ rather than exactly $50\%$. These detailed label distributions are further elaborated in Appendix \ref{sec:appendix_tie_rates}.
\subsection{Results and Analysis}

Current LLM judges still cannot reliably reproduce expert preferences in research ideation. Across the 14 judges in Table~\ref{tab:meta_eval}, Overall Quality Soft Accuracy ranges from 59.52\% to 72.56\%, with most judges scoring in the low-to-mid 60\% range. Feasibility has the lowest agreement, with only three judges exceeding 60\%. These results show that current LLM judges remain unreliable substitutes for expert preference labels on this benchmark.

We further measure judge order and length effects. In a 500-comparison A/B-swap analysis for each of 11 rerun judges, the original-order and swapped-order Overall Quality rankings have a Spearman correlation of $0.942$. Across comparable-strength battles, judges select the longer proposal in 50.9\%--62.0\% of decisive predictions across dimensions. Appendix~\ref{sec:appendix_judge_diagnostics} reports the setup and complete results.

\section{Conclusion}

Ideation Arena provides a dynamic and extensible expert-preference benchmark for evaluating research proposals before execution. Built on an expert-guided retrospective evaluation pipeline, Ideation Arena collects double-blind pairwise judgments from 105 active researchers, thereby establishing a human preference baseline for this inherently subjective task. From over 6,000 expert votes, we construct the Research Ideation Leaderboard, benchmarking 14 foundation models and 5 agent systems and revealing that agent architectures can substantially reshape base-model ideation quality under closed-context constraints. We also release Ideation Arena-Eval, a meta-evaluation benchmark for assessing automated judges. Results on Ideation Arena-Eval show that current LLM judges still cannot reliably reproduce proposal-stage expert preferences across the five evaluation dimensions.

\section*{Limitations}

The Research Ideation Leaderboard is inherently time-sensitive. The reported rankings reflect the performance of representative foundation models and scientific agent systems at the time of evaluation. As foundation models are frequently updated and agent frameworks continue to evolve, these rankings should not be viewed as permanent capability estimates. Instead, they provide a controlled snapshot of current systems under a unified evaluation protocol, motivating future updates of Ideation Arena to continuously track progress in automated research ideation. The main benchmark focuses on computer science. The research-agent comparison evaluates the reasoning and proposal-generation components retained under the shared closed-context setting.

\section*{Ethical Considerations}

Ideation Arena is intended to support the evaluation of automated research ideation systems rather than to replace human scientific judgment. A potential risk is that leaderboard scores may be over-interpreted as definitive measures of scientific creativity, although they reflect model behavior under a specific closed-context protocol and at a particular time. In addition, automated ideation systems may be misused to generate large volumes of superficially plausible but low-quality research proposals. We therefore emphasize that generated ideas should be treated as preliminary hypotheses that require expert scrutiny, methodological validation, and empirical verification before being used in real research workflows.

\section*{Acknowledgments}

This work is supported by Zhongguancun Academy (Grant No. C20250401), with additional support from the National Natural Science Foundation of China (Grant No. 23IAA02114).

\bibliography{references}

\clearpage
\appendix

\section{Supplementary Details and Analyses}

\subsection{Domain Taxonomy for Expert Profiling}
\label{sec:appendix_taxonomy}

We adopted the AAAI submission subject guidelines to construct the domain taxonomy used for expert profiling. The taxonomy contains eight primary categories:
\begin{itemize}[itemsep=1pt, topsep=2pt, parsep=0pt, leftmargin=*]
    \item Computer Vision
    \item Natural Language Processing
    \item Machine Learning Methods and Theory Reasoning
    \item Planning and Symbolic AI
    \item Data Mining and Big Data
    \item Robotics and Embodied AI
    \item Multi-agent Systems and Game Theory
    \item Interdisciplinary Applications and Social Impact
\end{itemize}
These categories are further subdivided into 53 secondary sub-fields for participant self-selection and research profiling.

\subsection{Evaluation Criteria for Human Review}
\label{sec:appendix_eval_criteria}

To ensure consistency in the double-blind review process, we asked expert annotators to apply the following criteria when comparing two proposals:

\begin{itemize}[itemsep=1pt, topsep=2pt, parsep=0pt, leftmargin=*]
    \item \textbf{Overall Quality:} Which research idea would you be more willing to invest time and resources in, with the goal of developing it into a potential paper or project?
    \item \textbf{Novelty:} Which response proposes a method with greater uniqueness and substantive depth, rather than relying on superficial combinations of buzzwords?
    \item \textbf{Feasibility:} Which response presents a more realistic experimental design that respects current hardware constraints and data availability, without assuming non-existent datasets or physically impossible mechanisms?
    \item \textbf{Significance:} Which response addresses a problem of greater importance or potential impact, rather than a trivial or marginal improvement?
    \item \textbf{Specificity:} Which response provides more concrete technical detail, such as specific loss functions, architectural modifications, or evaluation metrics, instead of remaining at a high level of abstraction?
\end{itemize}

\subsection{Expert Panel Composition and Research-Area Coverage}
\label{sec:appendix_expert_panel}

The expert panel includes PhD students, postdoctoral researchers, faculty members, and industry researchers. Table~\ref{tab:expert_primary_areas} reports self-reported primary-area expertise, and Table~\ref{tab:expert_secondary_interests} reports the five most frequent secondary research interests. Experts could report expertise in multiple areas, so the tables represent overlapping research-area coverage.

\begin{table}[t]
\centering
\small
\caption{Primary-area expertise among the 105 annotators.}
\label{tab:expert_primary_areas}
\begin{tabular}{@{}p{0.66\columnwidth}r@{}}
\toprule
\textbf{Primary area} & \textbf{Annotators} \\
\midrule
Natural Language Processing & 89 \\
Machine Learning Methods and Theory Reasoning & 85 \\
Interdisciplinary Applications and Social Impact & 75 \\
Computer Vision & 69 \\
Multi-agent Systems and Game Theory & 65 \\
Data Mining and Big Data & 63 \\
Robotics and Embodied AI & 49 \\
Planning and Symbolic AI & 16 \\
\bottomrule
\end{tabular}
\end{table}

\begin{table}[t]
\centering
\small
\caption{Most frequently reported secondary research interests.}
\label{tab:expert_secondary_interests}
\begin{tabular}{@{}p{0.66\columnwidth}r@{}}
\toprule
\textbf{Secondary research interest} & \textbf{Annotators} \\
\midrule
Large Language Models / LLMs & 81 \\
Prompt Engineering \& In-Context Learning & 58 \\
Multi-agent Systems & 53 \\
AI for Science & 51 \\
Deep Learning Architectures & 50 \\
\bottomrule
\end{tabular}
\end{table}

\subsection{Seed-Context Topic Distribution}
\label{sec:appendix_seed_distribution}

The final pool contains 2,191 unique seed-paper contexts. Each context is assigned one secondary-field label, which maps to one of the eight primary areas in Table~\ref{tab:seed_primary_distribution}. Table~\ref{tab:seed_secondary_distribution} lists the five largest secondary subfields. Percentages are rounded to one decimal place.

\begin{table}[t]
\centering
\small
\caption{Primary-area distribution of seed-paper contexts.}
\label{tab:seed_primary_distribution}
\begin{tabular}{p{0.62\columnwidth}rr}
\toprule
\textbf{Primary area} & \textbf{Count} & \textbf{\%} \\
\midrule
Natural Language Processing & 593 & 27.1 \\
Computer Vision & 497 & 22.7 \\
Machine Learning Methods and Theory Reasoning & 492 & 22.5 \\
Interdisciplinary Applications and Social Impact & 235 & 10.7 \\
Data Mining and Big Data & 137 & 6.3 \\
Planning and Symbolic AI & 118 & 5.4 \\
Robotics and Embodied AI & 73 & 3.3 \\
Multi-agent Systems and Game Theory & 46 & 2.1 \\
\bottomrule
\end{tabular}
\end{table}

\begin{table}[t]
\centering
\small
\caption{Five largest secondary subfields in the seed-context pool.}
\label{tab:seed_secondary_distribution}
\begin{tabular}{@{}rp{0.43\columnwidth}rr@{}}
\toprule
\textbf{Rank} & \textbf{Secondary subfield} & \textbf{Count} & \textbf{\%} \\
\midrule
1 & Large Language Models, LLMs & 290 & 13.2 \\
2 & Reinforcement Learning, RL & 180 & 8.2 \\
3 & Vision-Language Models / Multimodal & 171 & 7.8 \\
4 & Generative Vision Models & 168 & 7.7 \\
5 & Large Vision Models \& Foundation Models & 118 & 5.4 \\
\bottomrule
\end{tabular}
\end{table}

\subsection{Closed-Context Agent Adaptations}
\label{sec:appendix_closed_context}
\label{sec:appendix_agent_adaptations}

The agent leaderboard characterizes adapted workflows under the shared closed-context configuration. Table~\ref{tab:agent_adaptations} records the original workflow, the components retained for proposal generation, and the components removed or replaced in each evaluated variant.

\begin{table*}[t]
\centering
\scriptsize
\setlength{\tabcolsep}{3pt}
\caption{Components of the research-agent workflows evaluated under the shared closed-context protocol.}
\label{tab:agent_adaptations}
\begin{tabular}{p{0.12\textwidth}p{0.22\textwidth}p{0.25\textwidth}p{0.31\textwidth}}
\toprule
\textbf{Agent} & \textbf{Original workflow} & \textbf{Retained in the evaluated variant} & \textbf{Removed or replaced} \\
\midrule
AI-Researcher & Literature review, idea generation, algorithm design and implementation, validation and refinement, result analysis, and manuscript creation. & Multiple candidate directions, candidate selection, and structured proposal generation. & Literature and search tools, code implementation, experiment execution, result analysis, and manuscript writing. \\
\addlinespace
ResearchAgent & Expansion from a core scientific paper through publication and knowledge-entity retrieval, followed by iterative problem, method, and experiment-design refinement with reviewing agents. & Problem--method--experiment generation, validator agents, and iterative scoring. & Core-paper expansion, Semantic Scholar retrieval, knowledge-entity retrieval, and the external discovery loop. \\
\addlinespace
MOOSE-Chem & Inspiration retrieval, hypothesis composition, and hypothesis ranking from a research question, background survey, and inspiration corpus. & Inspiration screening, hypothesis generation, refinement and recombination, and hypothesis scoring. & Chemistry default corpus, TOMATO-Chem files, Web of Science or custom-corpus construction, and outside-knowledge exploration. \\
\addlinespace
Virtual Scientists & Team organization and collaborative idea generation through inter-team and intra-team discussion. & Team formation, discussion, idea generation, novelty checking, and abstract and proposal review. & AMiner/FAISS open retrieval, author and paper graph context acquisition, and broad multi-team search. \\
\addlinespace
SciMON & Literature-grounded idea generation with retrieval of past-paper inspirations and iterative novelty optimization. & Concept and inspiration extraction, hypothesis generation, and iterative novelty refinement. & Knowledge-graph, citation, and semantic-neighbor retrieval; T5 and biomedical training or evaluation; and external corpora. \\
\bottomrule
\end{tabular}
\end{table*}

\subsection{Proposal-Length Analysis}
\label{sec:appendix_length}

We first restrict the analysis to battles between systems whose anchored ratings differ by at most 100 points and measure how often the longer proposal is preferred. We then fit a length-adjusted Bradley--Terry model for each dimension:
\begin{equation}
P(i \succ j)=\sigma\!\left(s_i-s_j+\beta\bigl(\log L_i-\log L_j\bigr)\right),
\end{equation}
where $L_i$ and $L_j$ are the character counts of the final proposals shown to annotators. Tie and Both Bad outcomes are encoded as 0.5, matching the primary leaderboard construction. Table~\ref{tab:length_analysis} reports the observed longer-proposal selection rates and the Spearman correlations between the length-adjusted and original rankings.

\begin{table}[t]
\centering
\small
\caption{Proposal-length diagnostics by evaluation dimension.}
\label{tab:length_analysis}
\resizebox{\columnwidth}{!}{%
\begin{tabular}{lcc}
\toprule
\textbf{Dimension} & \makecell{\textbf{Longer proposal}\\\textbf{selected}} & \makecell{\textbf{Spearman vs.}\\\textbf{original ranking}} \\
\midrule
Overall Quality & 55.8\% & 0.8357 \\
Novelty & 52.8\% & 0.8478 \\
Feasibility & 56.5\% & 0.7930 \\
Significance & 55.6\% & 0.8835 \\
Specificity & 58.7\% & 0.7504 \\
\bottomrule
\end{tabular}%
}
\end{table}

\subsection{Seed-Proximity Diagnostic}
\label{sec:appendix_seed_proximity}

We use SPECTER2 to measure proximity between generated proposals and their corresponding seed papers. An LLM distills each seed paper into a proposal-style summary following the same structured template as the generated proposals. The mean similarity between a generated proposal and its seed-derived summary is 0.9346; the mean similarity between two model proposals generated for the same task is 0.9326. We normalize the shared within-task similarity using
\begin{equation}
\Delta(p)=\operatorname{sim}(p,\operatorname{seed}_t)
-\operatorname{mean}_{q\ne p,\,q\in t}\operatorname{sim}(p,q).
\end{equation}
A larger $\Delta(p)$ denotes greater seed-specific proximity after accounting for the similarity induced by the shared topic and literature context.

\begin{table}[t]
\centering
\small
\caption{Mean seed-proximity margins by leaderboard ranking group.}
\label{tab:seed_proximity_groups}
\begin{tabular}{lrr}
\toprule
\textbf{Ranking group} & \makecell{\textbf{Overall}\\\textbf{Mean $\Delta$}} & \makecell{\textbf{Novelty}\\\textbf{Mean $\Delta$}} \\
\midrule
Ranks 1--6 & $-0.0041$ & $-0.0089$ \\
Ranks 7--12 & $0.0017$ & $0.0057$ \\
Ranks 13--18 & $-0.0186$ & $-0.0186$ \\
Ranks 19--24 & $-0.0009$ & $-0.0009$ \\
\bottomrule
\end{tabular}
\end{table}

\begin{table}[t]
\centering
\small
\caption{Model-level association between mean seed-proximity margin and Elo rating.}
\label{tab:seed_proximity_correlations}
\begin{tabular}{lcc}
\toprule
\textbf{Elo dimension} & \textbf{Spearman $\rho$} & \textbf{95\% CI} \\
\midrule
Overall Quality & $-0.041$ & $[-0.281,\ 0.078]$ \\
Novelty & $-0.082$ & $[-0.311,\ 0.058]$ \\
\bottomrule
\end{tabular}
\end{table}

Across both rankings, the group means remain close to zero and show no monotonic increase with rank in this similarity-based diagnostic.

\subsection{Tie and Both Bad Sensitivity}
\label{sec:appendix_bothbad}

The primary Bradley--Terry analysis encodes both Tie and Both Bad as 0.5. In the sensitivity analysis, Tie remains a 0.5 outcome, while each Both Bad label is converted into two comparisons in which a virtual acceptable-quality anchor is preferred over each candidate proposal. Table~\ref{tab:bothbad_sensitivity} compares the resulting rankings with the original rankings. We also compute per-model Both Bad involvement rates and average them within groups defined by the original Overall Quality ranking (Table~\ref{tab:bothbad_rates}).

\begin{table}[t]
\centering
\small
\caption{Ranking sensitivity when Both Bad is modeled separately.}
\label{tab:bothbad_sensitivity}
\begin{tabular}{p{0.57\columnwidth}c}
\toprule
\textbf{Dimension} & \makecell{\textbf{Spearman vs.}\\\textbf{original}} \\
\midrule
Overall Quality & 0.9991 \\
Novelty & 0.9983 \\
Feasibility & 1.0000 \\
Significance & 1.0000 \\
Specificity & 1.0000 \\
\bottomrule
\end{tabular}
\end{table}

\begin{table}[t]
\centering
\small
\caption{Mean model-level Overall Quality Both Bad involvement rates by original ranking group.}
\label{tab:bothbad_rates}
\begin{tabular}{p{0.57\columnwidth}c}
\toprule
\textbf{Overall Quality ranking group} & \makecell{\textbf{Both Bad}\\\textbf{involvement}} \\
\midrule
Ranks 1--6 & 0.4\% \\
Ranks 7--12 & 1.6\% \\
Ranks 13--18 & 3.0\% \\
Ranks 19--24 & 4.5\% \\
\bottomrule
\end{tabular}
\end{table}

\subsection{Biology and Physics Pilot Details}
\label{sec:appendix_cross_domain}

Each preliminary pilot includes four domain experts, with each expert completing approximately 50 pairwise comparisons. Overlapping annotations are used to estimate pairwise agreement. Table~\ref{tab:cross_domain_agreement} reports agreement by dimension, and Table~\ref{tab:cross_domain_ci} adds 95\% bootstrap confidence intervals to the Overall Quality ratings reported in the main text.

\begin{table}[t]
\centering
\small
\caption{Pairwise agreement in the Biology and Physics pilots.}
\label{tab:cross_domain_agreement}
\begin{tabular}{lcc}
\toprule
\textbf{Dimension} & \textbf{Biology} & \textbf{Physics} \\
\midrule
Overall Quality & 0.543 & 0.542 \\
Novelty & 0.500 & 0.625 \\
Feasibility & 0.522 & 0.532 \\
Significance & 0.543 & 0.521 \\
Specificity & 0.587 & 0.562 \\
\midrule
Mean & 0.539 & 0.556 \\
\bottomrule
\end{tabular}
\end{table}

\begin{table}[t]
\centering
\scriptsize
\caption{Overall Quality ratings and 95\% bootstrap confidence intervals in the preliminary pilots.}
\label{tab:cross_domain_ci}
\resizebox{\columnwidth}{!}{%
\begin{tabular}{lcc}
\toprule
\textbf{Model} & \textbf{Biology: rating [95\% CI]} & \textbf{Physics: rating [95\% CI]} \\
\midrule
GPT-5.1 & 1237.6 $[1124.1,1367.1]$ & 1375.6 $[1236.9,1546.3]$ \\
AI-Researcher (DeepSeek V3.2) & 1165.5 $[1053.5,1286.3]$ & 1106.5 $[983.4,1236.6]$ \\
DeepSeek V3.2 & 854.8 $[738.6,965.6]$ & 916.1 $[792.8,1027.0]$ \\
GPT-4o & 742.1 $[612.5,856.7]$ & 601.8 $[424.6,743.9]$ \\
\bottomrule
\end{tabular}%
}
\end{table}

\subsection{LLM-Judge Order and Length Diagnostics}
\label{sec:appendix_judge_diagnostics}

We rerun the A/B-swap analysis on 500 comparisons for each of 11 judges and apply the analysis to all five evaluation dimensions. Gemini 3 Pro Preview, Grok 4, and Grok 4 Fast are excluded from this rerun because their original API endpoints were unavailable at the time of the experiment. For Overall Quality, judge rankings based on original-order and swapped-order Soft Accuracy have a Spearman correlation of $\rho=0.942$.

\begin{table}[t]
\centering
\scriptsize
\caption{Overall Quality Soft Accuracy and order-swap consistency.}
\label{tab:judge_order_swap}
\resizebox{\columnwidth}{!}{%
\begin{tabular}{lccc}
\toprule
\textbf{Judge} & \textbf{Original} & \textbf{Swapped} & \textbf{Consistency} \\
\midrule
Kimi K2 (Thinking) & 70.2\% & 70.4\% & 83.8\% \\
Claude 4.5 Opus & 69.5\% & 68.4\% & 88.1\% \\
DeepSeek V3.2 & 67.8\% & 67.2\% & 71.3\% \\
Qwen3-Max & 66.9\% & 67.2\% & 92.5\% \\
GLM-4.7 & 66.9\% & 69.0\% & 79.3\% \\
Mistral Large 2512 & 66.0\% & 66.2\% & 70.6\% \\
Claude 4.5 Sonnet & 64.6\% & 65.9\% & 90.4\% \\
GPT-5.1 & 64.5\% & 64.5\% & 86.6\% \\
Gemini 3 Flash Preview & 64.1\% & 63.7\% & 78.2\% \\
OpenAI o3 & 63.5\% & 63.4\% & 77.8\% \\
GPT-4o & 62.6\% & 60.9\% & 45.0\% \\
\bottomrule
\end{tabular}%
}
\end{table}

\begin{table*}[t]
\centering
\scriptsize
\setlength{\tabcolsep}{4pt}
\caption{Soft Accuracy averaged across the original and swapped presentation orders.}
\label{tab:judge_two_order_accuracy}
\begin{tabular}{lcccccc}
\toprule
\textbf{Judge} & \textbf{Overall Quality} & \textbf{Novelty} & \textbf{Feasibility} & \textbf{Significance} & \textbf{Specificity} & \textbf{Avg.} \\
\midrule
Kimi K2 (Thinking) & 70.3\% & 68.2\% & 59.9\% & 66.0\% & 66.4\% & 66.2\% \\
Claude 4.5 Opus & 68.9\% & 67.3\% & 60.8\% & 67.1\% & 69.8\% & 66.8\% \\
GLM-4.7 & 67.9\% & 65.2\% & 62.2\% & 61.9\% & 68.2\% & 65.1\% \\
DeepSeek V3.2 & 67.5\% & 65.8\% & 58.9\% & 64.8\% & 65.6\% & 64.5\% \\
Qwen3-Max & 67.1\% & 62.9\% & 60.5\% & 61.9\% & 65.8\% & 63.6\% \\
Mistral Large 2512 & 66.1\% & 61.7\% & 58.8\% & 64.3\% & 65.1\% & 63.2\% \\
Claude 4.5 Sonnet & 65.3\% & 62.2\% & 56.1\% & 61.7\% & 65.0\% & 62.1\% \\
GPT-5.1 & 64.5\% & 64.1\% & 58.4\% & 60.6\% & 64.2\% & 62.4\% \\
Gemini 3 Flash Preview & 63.9\% & 63.8\% & 56.5\% & 62.1\% & 64.4\% & 62.1\% \\
OpenAI o3 & 63.4\% & 63.6\% & 54.9\% & 59.2\% & 64.7\% & 61.2\% \\
GPT-4o & 61.7\% & 60.6\% & 56.0\% & 58.3\% & 60.4\% & 59.4\% \\
\bottomrule
\end{tabular}
\end{table*}

The ten judges other than GPT-4o have Overall Quality order-swap consistency between 70.6\% and 92.5\%; GPT-4o has 45.0\% consistency. Table~\ref{tab:judge_length_bias} reports the rate at which judges select the longer proposal among decisive predictions in comparable-strength battles.

\begin{table}[t]
\centering
\small
\caption{Longer-proposal selection rates for LLM judges.}
\label{tab:judge_length_bias}
\begin{tabular}{lc}
\toprule
\textbf{Dimension} & \textbf{Longer proposal selected} \\
\midrule
Overall Quality & 58.9\% \\
Novelty & 58.8\% \\
Feasibility & 50.9\% \\
Significance & 62.0\% \\
Specificity & 58.1\% \\
\bottomrule
\end{tabular}
\end{table}

\subsection{Elo Rating Confidence Intervals}
\label{sec:appendix_stats}

\begin{figure*}[t]
    \centering
    \includegraphics[width=0.99\textwidth]{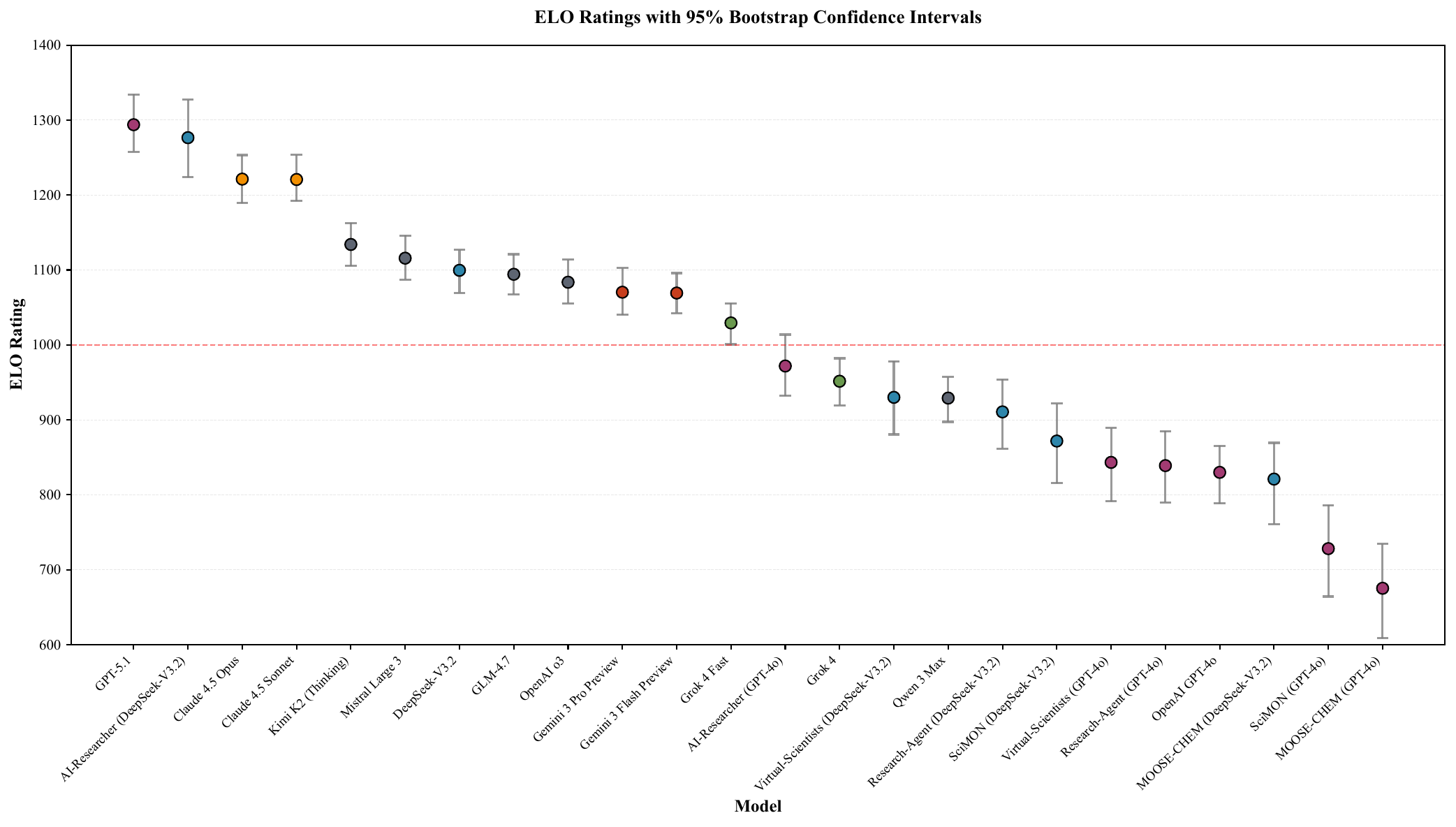}
    \caption{Elo Rating Confidence Intervals}
    \label{fig:elo_ci}
\end{figure*}

Here we provide detailed statistical evidence to support the leaderboard analysis in the main text.
Figure \ref{fig:elo_ci} presents the resulting leaderboard with error bars. A rigorous analysis of confidence intervals substantiates the statistical validity of our results on two critical fronts. The disjoint intervals between the AI-Researcher (DeepSeek V3.2) agent and its underlying base model confirm that the observed +177 point gain is a genuine improvement attributable to the agentic architecture, rather than an artifact of stochastic evaluation noise. This analysis corroborates the model stratification discussed in the main leaderboard results; while the top-tier models (GPT-5.1, AI-Researcher, and Claude 4.5 Opus) exhibit marginal intersection, their collective distribution is decisively separated from mid-tier open-source baselines, thereby empirically validating the existence of distinct performance tiers.

\subsection{Distribution of Human Preferences and Random Baselines}
\label{sec:appendix_tie_rates}

We present a detailed statistical analysis of the human preference annotations in the Ideation Arena-Eval benchmark. To contextualize the model performance metrics reported in the meta-evaluation experiments, we examine the marginal distributions of the ground truth labels---specifically across the five evaluation dimensions. In contrast to standard balanced classification tasks where a random guess yields 50\% accuracy, the existence of tie labels affects the expected baseline score. According to our evaluation protocol, a tie in the ground truth assigns 0.5 points to both the model and the baseline. Consequently, dimensions with a higher frequency of ties exhibit an inherently higher random baseline.

\begin{samepage}
\captionof{table}{Distribution of human expert preferences and corresponding Random Agreement baselines across five dimensions.}
\label{tab:human_pref_dist}
\vspace{2mm}
\begingroup
\small
\newcommand{\prefrow}[5]{%
    \noindent\makebox[0.36\linewidth][l]{#1}%
    \makebox[0.14\linewidth][c]{#2}%
    \makebox[0.14\linewidth][c]{#3}%
    \makebox[0.13\linewidth][c]{#4}%
    \makebox[0.20\linewidth][c]{#5}\par}
\noindent\rule{\linewidth}{0.7pt}\par
\prefrow{\textbf{Dimension}}{\textbf{A Win}}{\textbf{B Win}}{\textbf{Tie}}{\textbf{Random}}
\prefrow{}{}{}{}{\textbf{Agreement}}
\noindent\rule{\linewidth}{0.4pt}\par
\prefrow{D0: Overall Quality}{44.43\%}{44.95\%}{10.62\%}{50.57\%}
\prefrow{D1: Novelty}{42.03\%}{41.93\%}{16.04\%}{51.29\%}
\prefrow{D2: Feasibility}{40.80\%}{40.88\%}{18.32\%}{51.68\%}
\prefrow{D3: Significance}{40.40\%}{40.37\%}{19.24\%}{51.85\%}
\prefrow{D4: Specificity}{41.14\%}{44.06\%}{14.80\%}{51.14\%}
\noindent\rule{\linewidth}{0.7pt}\par
\endgroup
\end{samepage}
Table~\ref{tab:human_pref_dist} presents the empirical probabilities for each dimension. We observe that Significance displays the highest uncertainty among human experts, with a tie rate of $19.24\%$. This elevated ambiguity results in the highest random agreement baseline of $51.85\%$, suggesting that distinguishing the potential impact of two scientific ideas is inherently more subjective than assessing other metrics. In contrast, Overall Quality exhibits the lowest tie rate ($10.62\%$), indicating that experts form more distinct preferences when evaluating the general merit of a research proposal.

\subsection{Recruitment and Compensation}
Expert annotators were recruited through academic social-media platforms and screened based on their active research experience and self-reported research domains. Annotators received monetary compensation for their participation. The compensation was determined according to the expected annotation workload and was communicated to participants before annotation.

\section{Qualitative Case Study Analysis}
\label{sec:appendix_case_study}

\par\noindent
    \centering
    \begin{ideabox}{Case Study 1: High-Quality Response (Generated by GPT-5.1)}
    \section*{Title}
RELAX: Relational Elastic Adapters for Non-Forgetful Knowledge Injection in Large Language Models

\subsection*{1. Problem Formulation \& Motivation}

\subsubsection*{Context Synthesis}

The provided literature surfaces four interacting phenomena:

\begin{enumerate}
    \item \textbf{Knowledge injection vs retrieval.}
    \begin{itemize}
        \item ``Fine-Tuning or Retrieval?'' shows that unsupervised fine-tuning struggles to inject \textit{new} factual knowledge, even with multiple paraphrases, while RAG consistently outperforms parameter-only methods on knowledge-intensive tasks.
        \item ``Large Language Models Struggle to Learn Long-Tail Knowledge'' finds that parametric models need orders-of-magnitude more scale to capture rare facts, whereas retrieval-augmentation alleviates long-tail failures.
    \end{itemize}
    \item \textbf{Catastrophic forgetting in continual tuning.}
    \begin{itemize}
        \item Classical EWC-based continual learning (``Overcoming catastrophic forgetting in neural networks'') and RecAdam (``Recall and Learn'') show that regularization and pretraining simulations can mitigate forgetting.
        \item ``An Empirical Study of Catastrophic Forgetting in Large Language Models During Continual Fine-Tuning'' observes that LLMs (1B--14B) do forget, sometimes more severely as scale increases, and that general instruction tuning can partially buffer forgetting.
    \end{itemize}
    \item \textbf{Structured relational failures and factuality.}
    \begin{itemize}
        \item ``The Reversal Curse'' exposes a robust failure: training on ``A is B'' does not imply the model can answer the reverse ``B is A'', and naive data augmentation does not fix this.
        \item ``Language Models as Knowledge Bases?'', ``Crawling the Internal Knowledge-Base of Language Models'', and the factuality surveys reveal that while LMs encode many facts, they lack structured access and logical consistency (e.g., symmetry, inverses, compositions).
    \end{itemize}
    \item \textbf{Domain specialization \& knowledge-enhanced models.}
    \begin{itemize}
        \item Domain LLMs (Lawyer LLaMA, Med-PaLM, BloombergGPT, FinGPT) achieve strong performance via large-scale domain pretraining or heavy fine-tuning, which is expensive and brittle to further updates.
        \item Adapter-based methods (K-Adapter, adapter-based ConceptNet/OMCS injection, K-BERT) show that \textit{modular} knowledge infusion is possible without fully retraining the backbone, but they do not explicitly address: continual updates, long-tail acquisition, or relational constraints like the Reversal Curse.
    \end{itemize}
\end{enumerate}

\subsubsection*{The Critical Bottleneck}

Across these works, there is \textit{no} method that simultaneously provides:
\begin{itemize}
    \item \textbf{Local, continual knowledge updates} without catastrophic forgetting of general capabilities.
    \item \textbf{Relational consistency} for injected knowledge (e.g., symmetry, inverses) to overcome failures like the Reversal Curse.
    \item \textbf{Long-tail efficiency} --- capturing rare/new facts without massive-scale retraining.
\end{itemize}

\subsubsection*{Research Question}

Can we design a \textit{parameter-efficient, continual learning mechanism} for LLMs that:
\begin{enumerate}
    \item Injects new factual and domain knowledge \textit{modularly};
    \item Maintains prior performance on general and earlier tasks;
    \item Enforces \textit{relational structure} to overcome failures like the Reversal Curse;
    \item Efficiently handles long-tail facts by coupling parametric updates with retrieval?
\end{enumerate}

\subsection*{2. Proposed Methodology (The Core Contribution)}

\subsubsection*{Overview}

We propose \textbf{RELAX (Relational Elastic Adapters)}, a framework that:
\begin{itemize}
    \item Augments a frozen backbone LM with \textbf{relation-specific adapter modules} and \textbf{entity embeddings}.
    \item Trains using a \textbf{bidirectional relational objective} (coupling $(s, r, o)$ and $(o, r^{-1}, s)$).
    \item Uses \textbf{elastic regularization} (EWC + RecAdam-style KL anchoring) for continual injection.
    \item Integrates \textbf{retrieval} to bootstrap and distill knowledge into adapters.
\end{itemize}

\subsubsection*{2.1 Architecture}

Let $f_{\theta_0}$ be a pretrained LM. RELAX adds:
\begin{itemize}
    \item \textbf{Entity Memory $E$}: A dictionary mapping entity IDs to learned embeddings $e_e \in \mathbb{R}^d$.
    \item \textbf{Relational Adapters $A^{(\ell)}_r$}: Low-rank adapters defined as:
    $A^{(\ell)}_r(h) = W^{(\ell)}_{r,2} \, \sigma(W^{(\ell)}_{r,1} h)$. The updated hidden state is:
    $\tilde{h}^{(\ell)}(x) = h^{(\ell)}(x) + \sum_{r} g^{(\ell)}_r(x) \odot A^{(\ell)}_r(h^{(\ell)}(x))$.
    \item \textbf{Relational Scoring Head $f_r$}: Defined as $f_r(e_s, e_o) = w_r^\top \phi(e_s, e_o)$ for contrastive learning.
\end{itemize}

\subsubsection*{2.2 Training Objectives}

At continual phase $t$:
$\mathcal{L}^{(t)} = \mathcal{L}_{\text{task}}^{(t)} + \lambda_{\text{rel}} \mathcal{L}_{\text{rel}}^{(t)} + \mathcal{L}_{\text{elastic}}^{(t)}$

\begin{itemize}
    \item \textbf{Relational Loss $\mathcal{L}_{\text{rel}}$}: Includes fact scoring ($\mathcal{L}_{\text{rel-fwd}}$), inverse consistency ($\mathcal{L}_{\text{inv}}$), symmetry ($\mathcal{L}_{\text{sym}}$), and generative coupling for the Reversal Curse ($\mathcal{L}_{\text{gen-rev}}$).
    \item \textbf{Elastic Regularization $\mathcal{L}_{\text{elastic}}$}: Combines EWC-style consolidation and RecAdam-style KL anchoring:
    $\mathcal{L}_{\text{elastic}}^{(t)} = \lambda_{\text{EWC}} \sum_{i} F^{(t-1)}_i (\theta_i - \theta_i^{*(t-1)})^2 + \lambda_{\text{KL}} \text{KL}(p_{\theta^{*(t-1)}} \Vert p_{\theta})$.
\end{itemize}

\subsection*{3. Experimental Design Plan}

\begin{itemize}
    \item \textbf{Target Scenarios}: Reversal \& relational consistency; Continual domain specialization (Law $\to$ Med $\to$ Finance); Long-tail factual QA.
    \item \textbf{Baselines}: Vanilla Fine-tuning, LoRA, RAG, EWC, RecAdam, K-Adapter, and specialized LMs (Lawyer LLaMA, etc.).
    \item \textbf{Metrics}: Accuracy/F1, Forgetting measure, Reversal Gap, and Human eval for factuality/harmfulness.
\end{itemize}

\subsection*{4. Expected Contributions}

\begin{enumerate}
    \item \textbf{A unified framework} for modular, non-forgetful knowledge injection.
    \item \textbf{A relational objective} that explicitly addresses the Reversal Curse through shared parameterization.
    \item \textbf{Empirical demonstration} of efficient domain adaptation and long-tail coverage with high factual consistency.
\end{enumerate}
    \end{ideabox}
    \label{fig:gpt5.1_example}
\par

\par\noindent
    \centering
    \begin{ideabox}{Case Study 2: High-Quality Response (Generated by AI-Researcher with DeepSeek V3.2)}
    
    \section*{Title}
    Research Proposal: Compositional Multimodal Process Supervision via Dynamic Program Induction and Learned Verifiers (DynaProg)

    \subsection*{1. Challenges}

    \textbf{Current Technical Limitations:}
    \begin{enumerate}
        \item \textbf{Process Supervision Bottleneck in Multimodal Reasoning}: While process supervision has improved mathematical reasoning in language-only domains, its multimodal application is underdeveloped due to the scarcity of high-quality, stepwise annotated traces. Manual annotation is prohibitively expensive.
        \item \textbf{Lack of Compositional Reasoning Structure}: Current models treat reasoning as a linear chain. Real-world tasks (e.g., physics problems with diagrams) require hierarchical decomposition into sub-problems across different modalities.
        \item \textbf{Modality Alignment Verification Gap}: A reasoning step must be logically coherent \textbf{and} perceptually consistent with visual input. Current reward models lack explicit mechanisms to verify cross-modal grounding, leading to hallucinations.
        \item \textbf{Scalability of Self-Supervised Learning}: Generating high-quality, diverse reasoning programs that maintain logical validity without human intervention remains a substantial gap.
    \end{enumerate}

    \textbf{Unsolved Problems in Existing Work:}
    \begin{itemize}
        \item Absence of frameworks for automatic structured program induction from multimodal problems.
        \item Monolithic verification mechanisms lack specialization for distinct error types.
        \item Integration of program induction with process-supervised RL is unexplored in multimodal contexts.
    \end{itemize}

    \subsection*{2. Existing Methods}

    \begin{enumerate}
        \item \textbf{End-to-End Fine-Tuning}: Models like GPT-4V use CoT but lack explicit process guidance, making them prone to learning shortcuts.
        \item \textbf{Process-Supervised RL}: Relies on expensive human data or monolithic reward models that treat all reasoning steps equally.
        \item \textbf{Test-Time Search}: MCTS guided by rewards is computationally expensive and sample-inefficient.
        \item \textbf{Program Synthesis}: PAL/PoT translate reasoning to code but struggle with visual primitives and grounding continuous perception in discrete states.
    \end{enumerate}

    \subsection*{3. Motivation}

    Advancing multimodal reasoning requires moving from generating traces to \textbf{constructing and executing explicit reasoning programs}. This paradigm shift addresses the data bottleneck by automatically generating supervisory signals.

    \textbf{Potential Impact}: Accelerating scientific discovery, creating intelligent tutoring systems, and enabling auditable rationales in healthcare and autonomous systems.

    \subsection*{4. Proposed Method}

    \textbf{DynaProg Framework Overview}: DynaProg integrates dynamic program induction with modular verifiers and process-supervised RL.

    \subsubsection*{Stage 1: Unsupervised Program Induction}
    We define a Domain-Specific Language (DSL):
    \[
    \begin{aligned}
    \text{DSL} = \{&\text{READ}, \text{LOCATE}, \text{QUERY},\\
                  &\text{CALCULATE}, \text{COMPARE}, \text{IF},\\
                  &\text{WHILE}, \text{EXTRACT}\}
    \end{aligned}
    \]
    We select the most consistent program $P^*$ via clustering:
    \[ P^* = \arg\max_{P_i \in \mathcal{C}_{\text{majority}}} \sum_{j \neq i} \text{Sim}(P_i, P_j) \]

    \subsubsection*{Stage 2: Learning Modular Process Verifiers}
    Specialized verifiers $V_\phi = \{V_L, V_P, V_S\}$ evaluate steps:
    \begin{itemize}
        \item \textbf{Logical Verifier $V_L$}: $r_L^{(i)} = V_L(s_i \mid Q, s_{<i}, \mathcal{K})$
        \item \textbf{Perceptual Verifier $V_P$}: $r_P^{(i)} = V_P(s_i \mid I, \text{region}_i)$
        \item \textbf{Program-State Verifier $V_S$}: $r_S^{(i)} = V_S(s_i \mid \text{state}_{i-1}, \text{primitive}_i)$
    \end{itemize}
    Composite step reward: $r_i = \alpha r_L^{(i)} + \beta r_P^{(i)} + \gamma r_S^{(i)}$, where $\alpha + \beta + \gamma = 1$.

    \subsubsection*{Stage 3: RL with Composed Process Rewards}
    We train Actor $\pi_\theta$ via PPO with reward:
    \[ R(P, T) = \sum_{i=1}^{n} \gamma^{n-i} r_i + \lambda \cdot R_{\text{outcome}} \]

    \subsection*{5. Technical Details}

    \textbf{Algorithm 1: DynaProg Training Loop}
    \begin{enumerate}
        \item \textbf{Input:} Dataset $D$, policy $\pi_\theta$, verifiers $V_\phi$
        \item \textbf{For} epoch = 1 to $N$:
        \begin{itemize}
            \item Sample $(I, Q)$, generate $K$ programs $\{P_i\} \sim \pi_\theta$
            \item Cluster and select best program $P^*$ for $D_{\text{synth}}$
            \item Update $V_\phi$ using $D_{\text{synth}}$ via $\mathcal{L}_{V_L}, \mathcal{L}_{V_P}, \mathcal{L}_{V_S}$
            \item Update $\pi_\theta$ using PPO objective $\mathcal{L}^{\text{CLIP+KL}}(\theta)$
        \end{itemize}
        \item \textbf{End For}
    \end{enumerate}

    \subsection*{6. Expected Outcomes}

    \textbf{Anticipated Improvements}:
    \begin{itemize}
        \item \textbf{MathVista}: 8-12\% gain on vision-indispensable tasks.
        \item \textbf{ChartQA}: 12-18\% gain on complex interpretation.
        \item \textbf{Hallucination Rate}: 40-60\% reduction.
        \item \textbf{Data Efficiency}: 80-90\% reduction in human annotation needs.
    \end{itemize}

    \textbf{Long-Term Vision}: Laying the foundation for AI that solves complex problems while providing structured, auditable, and verifiable explanations.

    \end{ideabox}
\label{fig:gpt5.2_example}
\par

\par\noindent
    \centering
    \begin{ideabox}{Case Study 3: High-Quality Response (Generated by Claude 4.5 Opus)}
        \small % 使用稍小的字体以容纳大量内容
        \noindent\textbf{Title: Probabilistic Bidirectional Cross-Modal Embedding Alignment for Uncertainty-Aware Self-Supervised Terrain Segmentation}

        \vspace{0.5em}
        \noindent\textbf{1. Problem Formulation \& Motivation}
        
        \noindent\textbf{Context Synthesis:} Recent works have demonstrated the effectiveness of using proprioceptive sensors to self-supervise exteroceptive terrain classification. The ``Self-Supervised Visual Terrain Classification From Unsupervised Acoustic Feature Learning'' approach learns discriminative audio embeddings using triplets formed from visual feature similarity. Similarly, ``Where Should I Walk?'' uses force-torque signals, and ``A Self-Supervised Terrain Roughness Estimator'' uses accelerometer shock data. These methods follow a \textbf{unidirectional supervision} paradigm where one modality acts as the oracle for training the other.

        \noindent\textbf{The Critical Bottleneck:} The fundamental limitation is \textbf{error propagation without uncertainty quantification}:
        \begin{itemize}
            \item \textbf{Triplet Formation Noise:} Visual feature similarity used to form audio triplets is noisy due to illumination/domain shifts.
            \item \textbf{Clustering Uncertainty:} Hard cluster assignments are treated as ground truth despite uncertainty at boundaries.
            \item \textbf{Sparse Label Overconfidence:} All labeled pixels receive equal weighting regardless of confidence.
            \item \textbf{Temporal Misalignment:} Asynchronous signals are not explicitly modeled.
        \end{itemize}

        \noindent\textbf{Research Question:} \textbf{How can we design a framework that (1) enables bidirectional mutual regularization, (2) explicitly propagates uncertainty from embedding to segmentation, and (3) prevents low-confidence pseudo-labels from corrupting the classifier?}

        \vspace{0.5em}
        \noindent\textbf{2. Proposed Methodology (The Core Contribution)}

        \noindent\textbf{Overview:} We propose \textbf{Probabilistic Bidirectional Cross-Modal Alignment (PBCMA)}, replacing unidirectional supervision with a symmetric mutual embedding alignment in a shared probabilistic latent space.
    \end{ideabox}
\par

\par\noindent
    \centering
    \begin{ideabox}[notitle]{}
        \small

        \noindent\textbf{2.1 Probabilistic Embedding Networks:} Instead of point embeddings, we map samples to distributions:
        $f_a(a_i) \rightarrow (\mu_a^i, \Sigma_a^i)$ and $f_v(v_i) \rightarrow (\mu_v^i, \Sigma_v^i)$ (Mean and diagonal covariance in $\mathbb{R}^d$).

        \noindent\textbf{2.2 Bidirectional Cross-Modal Contrastive Loss:} A symmetric objective:
        \[
        \mathcal{L}_{BPCL} = \mathcal{L}_{a \rightarrow v} + \mathcal{L}_{v \rightarrow a} + \lambda_{KL} \cdot \mathcal{L}_{KL}
        \]
        where $\mathcal{L}_{a \rightarrow v}$ uses the 2-Wasserstein distance $D_W$ between distributions:
        \[
        \mathcal{L}_{a \rightarrow v} = -\frac{1}{N}\sum_{i=1}^{N} \log \frac{\exp(-D_{W}(p_a^i, p_v^i)/\tau)}{\sum_{j=1}^{N} \exp(-D_{W}(p_a^i, p_v^j)/\tau)}
        \]
        \[
        D_W(p_a^i, p_v^j) = ||\mu_a^i - \mu_v^j||_2^2 + ||\Sigma_a^{i,1/2} - \Sigma_v^{j,1/2}||_F^2
        \]

        \noindent\textbf{2.3 Consistency-Weighted Soft Clustering:} We use a GMM for soft cluster assignments $\gamma_{ik}$. The \textbf{cross-modal consistency score} measures agreement:
        \[
        c_i = 1 - \frac{1}{2}||(\gamma_a^i - \gamma_v^i)||_1
        \]

        \noindent\textbf{2.4 Uncertainty-Weighted Segmentation Loss:} Pixels are weighted by confidence:
        \[
        w_p = c_i \cdot h(\gamma_i) \cdot \exp(-\sigma_v^{i,2}/\beta)
        \]
        The segmentation network is trained with weighted soft cross-entropy: $\mathcal{L}_{seg} = -\sum_{p \in \mathcal{T}} w_p \sum_{k=1}^{K} \hat{y}_{p,k} \log q_{p,k}$.

        \noindent\textbf{2.5 Temporal Consistency Regularization:}
        \[
        \begin{aligned}
        \mathcal{L}_{temp}
        ={}& \sum_{t} D_W(p_a^t, p_a^{t+1})
        + D_W(p_v^t, p_v^{t+1}) \\
        &- \alpha \cdot \mathbb{1}[\text{terrain\_change}_t]
        \end{aligned}
        \]

        \noindent\textbf{Why It Works:} 1) Bidirectional learning prevents error accumulation. 2) Probabilistic embeddings capture uncertainty. 3) Cross-modal consistency identifies reliable samples. 4) Wasserstein loss handles multimodal distributions robustly.

        \vspace{0.5em}
        \noindent\textbf{3. Experimental Design Plan}
        \begin{itemize}
            \item \textbf{Target:} Pixelwise terrain classification (e.g., asphalt, grass) for mobile robots.
            \item \textbf{Baselines:} AudioVisual-UATC (Unidirectional), LSTM-AudioCNN, Force-Torque Self-Supervision, Fully Supervised Upper Bound, DeepLabv3+ (standard pseudo-labels).
            \item \textbf{Metrics:} mIoU, Expected Calibration Error (ECE), Label Efficiency, Robustness (SNR/Lighting), Inference FPS.
            \item \textbf{Datasets:} Freiburg Forest/Crossing, Indoor Multi-Terrain, RUGD.
        \end{itemize}

        \vspace{0.5em}
        \noindent\textbf{4. Expected Contributions}
        \begin{enumerate}
            \item \textbf{We propose PBCMA}, a novel framework replacing unidirectional supervision with symmetric probabilistic embedding alignment.
            \item \textbf{We introduce a principled uncertainty propagation mechanism} where cross-modal consistency prevents low-confidence pseudo-labels from corrupting the visual classifier.
            \item \textbf{We demonstrate empirically} that PBCMA achieves comparable mIoU to fully supervised methods while requiring zero manual labels, outperforming unidirectional baselines by 8-12\% mIoU.
        \end{enumerate}
    \end{ideabox}

\par\noindent
    \centering
    \begin{badbox}{Case Study 4: Low-Quality Response (Brevity)}
    \subsection*{1. MOOSE-Chem 2026 (DeepSeek V3.2)}
    We hypothesize that a **Hierarchical Coordinated Prompt Tuning (HCPT)** method can efficiently adapt large frozen vision-language models (e.g., CLIP) to downstream tasks. The core novel mechanism is the learning of a two-tiered prompt structure: (1) a small set of **Global Coordination Prompts (GCPs)**, shared across both modalities, and (2) two separate sets of **Modality-Specific Prompts (MSPs)** for vision and text. These prompts are dynamically integrated via a **Coordinated Fusion Module (CFM)**, a lightweight, trainable network that conditions the prompt composition on the input modality."
    \subsection*{2. MOOSE-Chem 2026 (GPT-4o)}
    starts:** We hypothesize that a unified, general-purpose vision model for both image and video tasks can be achieved by integrating a meta-learned conditional codebook for efficient multi-task representation with a temporally coherent tokenization and decoding mechanism for video. The core innovation is a dual-path architecture where a meta-learned hypernetwork dynamically adapts a shared codebook for specific tasks, while a temporally-aware tokenization and a causal, uncertainty-aware decoder ensure stable and consistent outputs for sequential data.
    \subsection*{3. SciMON 2026 (GPT-4o)}
    **Title:** Event Extraction through Interactive Agent Feedback: A Reinforcement Learning Approach

**Motivation:** Current event extraction methodologies often rely on static structures, either treating the task as graph-based or question-answering based, both of which may fail to capture the dynamic nature of natural text. There is a clear need for a more adaptive framework that simulates human-like interactions and prioritizations, enabling more nuanced and contextually enriched event extraction. This research addresses the critical gap of adaptability and contextual understanding in event extraction tasks.

**Core Method/Architecture:** The proposed framework, Event Extraction through Interactive Agent Feedback (EEIAF), utilizes a multi-agent system where each agent simulates a different reader persona, focusing on specific aspects such as causality, relevance, or temporality. These agents are trained using reinforcement learning algorithms, such as Proximal Policy Optimization (PPO), to provide feedback to a central extraction module. The system dynamically adjusts the extraction process based on feedback loops, which are informed by each agent's unique focus area. The architecture comprises a central controller that aggregates agent feedback to refine event extraction decisions without relying on predefined structural constraints, thereby enabling the system to adaptively prioritize different text features according to context.

**Experimental Plan:** The experimental evaluation will be conducted using diverse datasets, such as ACE 2005 and the Rich ERE (Event and Relation Extraction) dataset, to ensure robustness across different domains. Baseline methods will include graph-based and QA-based event extraction techniques. Evaluation metrics will involve precision, recall, F1-score, and a novel adaptability index that measures the system's responsiveness to context variations. We expect EEIAF to outperform baseline methods in terms of both traditional metrics and adaptability, demonstrating enhanced event understanding in dynamic environments.

**Expected Contributions:** This research introduces a novel reinforcement learning-based architecture for event extraction that leverages interactive agent feedback, offering significant improvements in adaptability and contextual understanding. The EEIAF framework presents a paradigm shift from static to dynamic extraction processes, potentially setting new standards for event extraction tasks and influencing future research directions in natural language processing.
    \end{badbox}
    
    \label{fig:bad_breif_example}
\par

\par\noindent
    \centering
    \begin{badbox}{Case Study 5: Low-Quality Response (Conceptual Hollowness)}
    \subsection*{1. SciMON 2026 (GPT-4o)}
    \textbf{Title:} Enhancing Multi-Agent Communication with Synergistic Integration of Pre-trained Language Models and Graph Neural Networks

\textbf{Motivation:} Current multi-agent communication systems often fall short in leveraging deep contextual understanding while reasoning about complex relational interactions. This research addresses the critical gap by combining transformer-based language models and graph neural networks (GNNs) to enhance contextual understanding and relational reasoning. This work is vital for improving coordination and decision-making in multi-agent environments, such as autonomous vehicle fleets and robotic swarms, where sophisticated and contextually aware communication protocols are paramount.

\textbf{Core Method/Architecture:} Our approach introduces a novel framework that integrates large-scale pre-trained language models with graph neural networks to create a unified communication protocol. The language models, such as transformers, are utilized to capture nuanced contextual information from agent communications, leveraging their ability to understand and generate human-like text. Simultaneously, GNNs are employed to model the relational dynamics between agents, capturing the topological structure and interactions within the communication network. The architecture involves a dual-pathway system where the language model processes incoming messages to extract and encode context, which is then fed into the GNN to refine relational reasoning and update the communication protocol dynamically. The combined architecture is trained end-to-end using reinforcement learning techniques to optimize coordination and decision-making.

\textbf{Experimental Plan:} We will evaluate the proposed framework on simulated multi-agent environments, such as autonomous vehicle coordination tasks and robotic swarm scenarios. Baseline methods for comparison will include standalone transformer-based communication models and GNN-centric approaches. Evaluation metrics will focus on coordination efficiency, decision-making accuracy, and communication overhead. We expect our model to demonstrate superior performance in terms of these metrics, showcasing enhanced coordination and decision-making capabilities due to the synergistic integration of deep contextual understanding and relational reasoning.

\textbf{Expected Contributions:} This research is expected to deliver a novel integrated framework for multi-agent communication systems, showcasing substantial advancements in contextual and relational reasoning capabilities. The proposed method will significantly impact applications in autonomous systems and robotics, providing a robust foundation for developing more efficient, contextually aware communication protocols. Moreover, the methodological insights from this work could inspire further advancements in combining language understanding with structured relational modeling in other domains.

    \subsection*{2. MOOSE-Chem 2026 (GPT-4o)}
    To advance the evaluation of large language models (LLMs), we propose the Dynamic Multi-Modal Adaptive Benchmarking (DMAB) framework. This framework synthesizes adaptive benchmarking, multi-modal evaluation, and user interaction data integration to provide a robust, comprehensive, and scalable assessment of LLM performance.

1. Dynamic Query Synthesis and Integration: DMAB continuously gathers queries from diverse sources, including social media, forums, and user-facing platforms. Using advanced NLP techniques and semantic embeddings, such as those from BERT or Sentence Transformers, it selects queries that reflect diverse and current real-world contexts. Dynamic query synthesis is enhanced through generative models to create queries targeting model weaknesses.

2. Adaptive Weighting and Feedback Mechanism: Each query is dynamically weighted based on real-world frequency and performance variance across LLMs, using a reinforcement learning model to optimize weighting strategies. A feedback loop integrates user satisfaction and performance metrics to inform query synthesis and weight adjustments, ensuring alignment with real-world application demands.

3. Multi-Modal Evaluation Framework: Incorporate text, image, audio, and video data into the evaluation process. Utilize pre-trained models for each modality and implement a fusion strategy to combine modality-specific scores, providing a comprehensive assessment of LLM capabilities. This approach ensures that LLM evaluations reflect diverse content types and interaction modes.

4. Crowd-Sourced and Automated Grading Fusion: Select a subset of queries for human evaluation using crowd-sourced platforms, ensuring diversity in demographics and expertise. Employ Bayesian calibration to align machine evaluations with human judgments, using a weighted scoring mechanism to balance automated and human feedback.

5. Scalable, Cost-Effective Infrastructure: Implement a cloud-based, distributed processing architecture using services like AWS Lambda and Kubernetes. This infrastructure supports scalable and efficient data processing, enabling rapid updates and dynamic evaluation cycles while minimizing cost.

By integrating these components, DMAB aims to provide a precise, adaptable, and realistic evaluation of LLMs, capable of evolving with technological advancements and user expectations.

    \end{badbox}
    \label{bad_Conceptual_Hollowness_example}
\par

\par\noindent
    \centering
    \begin{badbox}{Case Study 6: Low-Quality Response (Structural Deficiency)}
    \subsection*{1. MOOSE-Chem 2026 (GPT-4o)}
    We propose an advanced framework for querying knowledge graphs (KGs) that integrates vector-based processing, multimodal embeddings, and iterative refinement through user feedback. This framework addresses challenges such as incomplete KGs, semantic matching, and natural language questions (NLQs) by combining the most effective components from existing hypotheses.

1. Multimodal KG Embedding Module: We introduce a unified embedding space for text, image, and audio data. This is achieved by employing models like BERT for text, CLIP for images, and Wav2Vec for audio. A gated attention mechanism dynamically integrates these embeddings based on query relevance, ensuring each modality is appropriately weighted.

2. Transformer-based Query Processing: A customized transformer model, leveraging BERT fine-tuned for multimodal input, is employed to process both structured queries and NLQs. This model incorporates additional layers to handle multimodal data and is trained using supervised contrastive learning on diverse datasets, enhancing its ability to capture complex semantic relationships.

3. Vector-based Query Matching and Inference: A novel query matching algorithm utilizes cosine similarity for vector relevance measurement, enhanced by a re-ranking mechanism based on semantic similarity scores. To address incomplete KGs, we implement a graph neural network-based predictive component using relational graph convolutional networks (R-GCNs) to infer missing triples, with confidence scores guiding the reliability of predictions.

4. Unified Query Processing Engine: The framework includes a query processing engine capable of handling various query types. It employs approximate nearest neighbor search techniques for efficient similarity searches and integrates cross-modal attention to maintain semantic coherence.

5. Iterative Refinement with User Feedback: User interactions are logged and analyzed through reinforcement learning, adjusting model parameters and embeddings based on reward-based signals. This feedback loop ensures continuous improvement in query accuracy, with metrics such as user satisfaction and query success rates guiding iterative refinement.

This framework offers a comprehensive approach to KG querying, integrating multimodal data and advanced machine learning techniques to enhance semantic understanding and query completeness.
    \subsection*{2. SciMON 2026 (GPT-4o)}
    \textbf{Title:} Multi-Modal Temporal Ensembling for Climate-Impact Narration (MOTECIN): Bridging Climate Forecasts and Socio-Economic Decision-Making

\textbf{Motivation:} Traditional climate models primarily focus on enhancing the accuracy of meteorological predictions without adequately addressing how these forecasts translate into socio-economic impacts. This creates a gap in actionable insights for non-expert decision-makers, such as city planners and local businesses, who require comprehensible narratives to prepare for and respond to climate events. MOTECIN aims to fill this gap by providing detailed, data-driven narratives that link predicted weather events directly to socio-economic outcomes.

\textbf{Core Method/Architecture:} MOTECIN proposes a novel multi-modal architecture that combines a Temporal Convolutional Network (TCN) with attention mechanisms and bidirectional LSTM networks to process both structured meteorological data and unstructured socio-economic data. The TCN handles temporal patterns from climate forecasts, while the attention-enhanced LSTM processes and synthesizes information from social media, local economic indicators, and news articles. A GAN-based downscaling module is employed to refine the spatial resolution of weather data, inspired by state-of-the-art image super-resolution techniques. The final output layer translates these multi-modal inputs into coherent narratives, which are optimized using a dual-objective function that balances predictive accuracy and narrative coherence.

\textbf{Experimental Plan:} The experimental evaluation will utilize datasets from the European Centre for Medium-Range Weather Forecasts (ECMWF) for meteorological data, alongside socio-economic datasets from sources such as the World Bank and Twitter sentiment analysis datasets. Baseline methods for comparison will include traditional downscaling approaches and recent deep learning models for precipitation forecasting. Evaluation metrics will focus on both quantitative measures, such as RMSE and F1 score for predictive accuracy, and qualitative assessments, such as readability and relevance of generated narratives as evaluated by domain experts. We expect MOTECIN to outperform baselines in both predictive accuracy and the utility of narratives.

\textbf{Expected Contributions:} MOTECIN is expected to advance the field of climate impact prediction by introducing an innovative method for converting complex climate data into actionable socio-economic insights. This research will enhance the interpretability and accessibility of climate forecasts for non-expert users, directly supporting community resilience and strategic planning. Additionally, MOTECIN's integration of multi-modal data represents a significant technical contribution to machine learning methodologies in the context of environmental science.
    \end{badbox}
    \label{bad_Structural_Deficiency_example}
\par

\newpage
\section{Prompt Templates}
\label{sec:prompt}
\par\noindent
    \centering
    \begin{promptbox}{Prompt template for LLM-generated research ideas}
    \section*{System Role} You are a Principal Researcher at a top-tier Computer Science research laboratory (e.g., MIT CSAIL, Berkeley EECS, Microsoft Research, or Google Research). You are brainstorming a new \textbf{methodology-focused} paper submission for a leading conference in the field.

\section*{Context: Literature Review} The following is a JSON array of specific research papers acting as your background reading. \begin{verbatim} <papers> {{JSON_INPUT_STRING}} </papers> \end{verbatim}

\section*{Task} Based on the provided papers, devise a \textbf{concrete, technical, and novel algorithmic, system-design, or theoretical solution} to an unresolved technical problem. The goal is to propose a ``Method Research Idea'' that introduces a new mechanism, protocol, algorithm, or system framework.

\section*{Constraints \& Guidelines} \begin{enumerate} \item \textbf{NO Reviews/Surveys:} Do not propose a literature survey, landscape analysis, or position paper.'' \item \textbf{NO Pure Benchmarks/Datasets:} Do not propose the creation of a new dataset, benchmark suite, or a simple comparative study as the \textit{primary} contribution. The contribution must be a \textit{method} (e.g., a new scheduling algorithm, a novel consistency protocol, a compiler optimization) that solves a problem better. \item \textbf{Technical Depth:} Do not use buzzwords without substance. Speak in specific technical terms relevant to the sub-field (e.g., distributed consensus,'' memory consistency models,'' cryptographic primitives,'' compiler intermediate representations,'' index structures,'' or ``algorithmic complexity''). \item \textbf{Logic Flow:} Your proposal must derive logically from the input papers. Explicitly reference how your idea improves upon or combines the specific techniques mentioned in the input. \end{enumerate}

\section*{Output Format} Please strictly follow the structure below. Do not output conversational filler.

\subsection*{Title} [A specific, technical title. Example: Zero-Copy Data Transfer via Speculative Execution'' instead of Better System Performance'']

\subsection*{1. Problem Formulation \& Motivation} \begin{itemize} \item \textbf{Context Synthesis:} How do the input papers currently approach the problem? (Mention specific methods/systems from input). \item \textbf{The Critical Bottleneck:} What is the specific technical limitation in these existing works? (e.g., High tail latency,'' Write amplification,'' Scalability limits in distributed settings,'' Vulnerability to side-channel attacks,'' ``NP-hard verification complexity''). \item \textbf{Research Question:} Formulate the precise engineering or mathematical question this project aims to answer. \end{itemize}

\subsection*{2. Proposed Methodology (The Core Contribution)} \begin{itemize} \item \textbf{Overview:} A high-level summary of your proposed framework or system. \item \textbf{Key Technical Innovation:} Detail the specific mechanism. \begin{itemize} \item \textit{If Algorithmic/Theoretical:} Describe the new algorithm, proof technique, or optimization bounds. \item \textit{If System/Architectural:} Describe the new system design, hardware-software interface, or module interaction. \item \textit{If Data/Storage:} Describe the new indexing structure, encoding scheme, or storage format (NOT just collecting new data). \end{itemize} \item \textbf{Why it Works:} Briefly explain the intuition behind why this modification solves the bottleneck identified above. \end{itemize}

\subsection*{3. Experimental Design Plan} \begin{itemize} \item \textbf{Target Task/Scenario:} What specific application or workload is this applied to? (e.g., Key-Value Store, Network Packet Filtering, Program Synthesis, Image Classification). \item \textbf{Baselines:} Which existing methods/systems (ideally from the input papers) would serve as the primary comparison? \item \textbf{Evaluation Metrics:} What specific metrics will determine success? (e.g., Throughput (RPS),'' p99 Latency,'' Proof Verification Time,'' Energy Consumption,'' ``False Positive Rate''). \end{itemize}

\subsection*{4. Expected Contributions} \begin{itemize} \item Summarize the 3 main contributions of this work (e.g., ``1. We propose [System Name], a novel approach for... 2. We prove that... 3. We demonstrate a 20\% reduction in overhead compared to...''). \end{itemize}
    \end{promptbox}
    \label{fig:llm_prompt}
\par

\par\noindent
    \centering
    \begin{promptbox}{LLM-as-a-Judge Prompt Template for Ideation Arena-Eval}
    \noindent You are an expert academic research reviewer. Please carefully read the research query and two research ideas (Response A and Response B) below, then evaluate which one is better across multiple dimensions.

\vspace{1em}

\noindent \textbf{Response A:} \\
\{response\_a\}

\noindent\rule{\linewidth}{0.4pt}

\noindent \textbf{Response B:} \\
\{response\_b\}

\noindent\rule{\linewidth}{0.4pt}

\vspace{1em}

\noindent Please evaluate both responses across the following five dimensions:

\begin{description}
    \item[D1. Novelty] \hfill \\
    \textbf{Question:} Which response proposes a more unique approach with substantial depth that goes beyond simple buzzword stacking, offering a fresh perspective or unexpected angle to the problem? \\
    \textbf{Your judgment:} [A/B/Tie]

    \item[D2. Feasibility] \hfill \\
    \textbf{Question:} Which response demonstrates a more realistic experimental design that fully considers current hardware limitations and data availability, avoiding hallucinations about non-existent datasets or physically impossible mechanisms? \\
    \textbf{Your judgment:} [A/B/Tie]

    \item[D3. Scientific Significance] \hfill \\
    \textbf{Question:} Which response addresses a more important or valuable problem, clearly establishing the potential impact and relevance of the proposed research to academia, rather than focusing on trivial improvements or marginal issues? \\
    \textbf{Your judgment:} [A/B/Tie]

    \item[D4. Specificity] \hfill \\
    \textbf{Question:} Which response provides more concrete technical details (e.g., specific loss functions, architecture modifications, or evaluation metrics) rather than staying at high-level abstractions or using vague terminology? \\
    \textbf{Your judgment:} [A/B/Tie]

    \item[D0. Overall Review Utility] \hfill \\
    \textbf{Question:} From a researcher's perspective, which research idea would you prefer to invest your time and resources in as a potential paper or project? \\
    \textbf{Your judgment:} [A/B/Tie]
\end{description}

\noindent\rule{\linewidth}{0.4pt}

\noindent Please provide your evaluation in the following format:

\begin{verbatim}
**D0_Overall:** [A/B/Tie]
**D1_Novelty:** [A/B/Tie]
**D2_Feasibility:** [A/B/Tie]
**D3_Significance:** [A/B/Tie]
**D4_Specificity:** [A/B/Tie]
\end{verbatim}

\noindent \textbf{Note:}
\begin{itemize}
    \item Choose ``A'' if Response A is clearly better
    \item Choose ``B'' if Response B is clearly better
    \item Choose ``Tie'' if both responses are comparable in quality
\end{itemize}

\noindent Don't provide any other text or explanation. Just return the evaluation in the format above.
    \end{promptbox}
    \label{fig:eval_prompt}
\par

\end{document}